\documentclass[12pt]{article}
\usepackage[utf8]{inputenc}
\usepackage{xcolor, times, soul}
\usepackage{multirow}
\usepackage{subfig}
\usepackage{graphicx}
\usepackage{float}
\usepackage{todonotes}
\usepackage{longtable}
\usepackage{colortbl, tcolorbox}
\usepackage{adjustbox,setspace}
\usepackage[labelfont=bf]{caption}
\usepackage[top=1in, bottom=1in, left=1in, right=1in]{geometry}

\usepackage[colorlinks=true, linkcolor=blue, citecolor=blue]{hyperref} 

\usepackage[backend=bibtex,style=nature,sorting=none, maxbibnames=99, natbib]{biblatex} 

\title{Benefits and Harms of Large Language Models \\in Digital Mental Health}

\author{Munmun De Choudhury$^\dagger$, Sachin R. Pendse, Neha Kumar\\
\normalsize{School of Interactive Computing, Georgia Institute of Technology, Atlanta, GA, USA}\\
\normalsize{\{munmund, sachin.r.pendse, neha.kumar\}@gatech.edu}\\
}

\date{}

\begin{document}

\maketitle

\small{$\dagger$ 
Corresponding author
}

\begin{abstract}
The past decade has been transformative for mental health research and practice. The ability to harness large repositories of data, whether from electronic health records (EHR), mobile devices, or social media, has revealed a potential for valuable insights into patient experiences, promising early, proactive interventions, as well as personalized treatment plans. Recent developments in generative artificial intelligence, particularly large language models (LLMs), show promise in leading digital mental health to uncharted territory. Patients are arriving at doctors' appointments with information sourced from chatbots, state-of-the-art LLMs are being incorporated in medical software and EHR systems, and chatbots from an ever-increasing number of startups promise to serve as AI companions, friends, and partners. This article presents contemporary perspectives on the opportunities and risks posed by LLMs in the design, development, and implementation of digital mental health tools. We adopt an ecological framework and draw on the affordances offered by LLMs to discuss four application areas---care-seeking behaviors from individuals in need of care, community care provision, institutional and medical care provision, and larger care ecologies at the societal level. We engage in a thoughtful consideration of whether and how LLM-based technologies could or should be employed for enhancing mental health. The benefits and harms our article surfaces could serve to help shape future research, advocacy, and regulatory efforts focused on creating more responsible, user-friendly, equitable, and secure LLM-based tools for mental health treatment and intervention.

\end{abstract}

\section{Introduction}
\begin{quote}
{\it ``When I use ChatGPT to talk things through and vent about how I feel, it goes on to tell me to get help and that I'm not alone. But why does it feel as if it's mocking me? It feels as if it's having a laugh at my expense.''} -- A paraphrased social media post
\end{quote}

In November 2022~\cite{openai2022chatgpt}, OpenAI released ChatGPT. ChatGPT followed the mold of past chatbots by providing a simple interface for people to easily interact with a conversational agent. However, unlike past publicly accessible chatbots, ChatGPT was powered by OpenAI's proprietary language generation model, often called Large Language Models (LLMs). OpenAI's LLM (named GPT, for Generative Pre-trained Transformers~\cite{radford2018improving}) was created through a large-scale collection of text from the Internet combined with manual review through a process often called Reinforcement Learning From Human Feedback (RLHF)~\cite{ziegler2019fine}. ChatGPT's underlying language and simple interface astonished users with answers that were surprisingly coherent and wide-ranging. Since then, conversations across academic, medical, industry, and policy domains have begun to discuss how LLMs could offer new possibilities for diagnosis, treatment, and patient care in mental health.  

Over the past decade, there has been increased conversation around the growing potential for digital technologies, artificial intelligence (AI), and machine learning to contribute value to mental health research and practice. Research has demonstrated some of this potential. For example, methods from natural language processing (such as sentiment analysis) have been used to assess people's emotional states from their text, speech, or social media language~\cite{moura2022digital, chancellor2020methods}. These studies have consistently shown that computational or predictive analyses of digital data can accurately detect mood~\cite{de2013predictingPPDChanges}, mental health states~\cite{de2013predicting}, 
and even the risk of potential harm~\cite{de2016discovering} and suicide~\cite{choi2020development}. Collectively, the implications of this research include the potential for valuable insights into daily patient experiences, a paving of the way for early and proactive interventions, and the design of personalized treatment plans. However, as people further rely on online tools to seek care for their mental health, researchers and activists have sounded the alarm about the potential for harm if digital mental health interventions are staged without the awareness or consent of people experiencing distress~\cite{pendse2022treatment, bossewitch2022digital}. Scholars have also expressed concern that the use of predictive analytics in mental health could compromise patient and clinician agency~\cite{chancellor2019human}, exacerbate systemic disparities encoded in the training data of AI models~\cite{thieme2020machine,pendse2022treatment}, and propagate insights poor in clinical grounding or construct validity~\cite{ernala2017linguistic}. The challenges in implementing an AI-informed mental health care model have further invited criticism and skepticism around the role of AI in this field~\cite{koutsouleris2022promise}, even as researchers have increasingly sought to draw upon both computational and psychiatric expertise and paradigms to advocate for an ``(AI) model-based psychiatry''~\cite{hauser2022promise}.   
With the rapid introduction of LLMs in new parts of everyday life, initiatives in digital mental health are likely to be similarly disrupted, opening doors to new opportunities for mental healthcare, while also setting the stage for new and previously unconsidered harms. 

This discussion of benefits and harms can already be seen unfolding in popular discourse around LLMs and mental health. For instance, ChatGPT was not created as a mental health support tool---however, people in distress have started to use ChatGPT for mental health support and non-judgmental guidance, as the opening quote to this Introduction shows. As discussed in online communities (such as Reddit~\cite{reardon2023chatbots}), users have turned to ChatGPT in moments of suicidal ideation, and expressed their belief that ChatGPT saved their life in moments of despair. Other users have described their approach to carefully training LLM-based chatbots to behave as therapists from different theoretical orientations, such as prompting chatbots to take on the role of an Acceptance and Commitment (ACT) therapist. Several mental healthcare organizations and companies~\cite{replika2023, crasto2021, oleary2023} have also begun to research the integration of LLMs into the design of their services. This increased use of LLMs in service delivery has been met with excitement~\cite{ayers2023evaluating}, but also with justified skepticism, given potential racial or gender biases~\cite{zack2023coding, omiye2023beyond} and unexpected outputs~\cite{jargon2023} from LLM-based chatbots. For example, in June 2023, the National Eating Disorder Association was forced to shut down a chatbot created to provide clinically validated information after the chatbot provided harmful and dangerous advice to users, including diet and weight loss advice~\cite{xiangtessa2023, jargon2023}. Unbeknownst to the administrative staff at NEDA, the chatbot company that provided its services to NEDA had ``rolled out an AI component to its chatbots''~\cite{jargon2023}. The chatbot was introduced soon after several pivotal events, including an attempt by NEDA helpline volunteers to unionize and the closing of the NEDA helpline, raising awareness of the potential that nascent technology may be used to replace human staff~\cite{xiangtessa2023}. The harms of irresponsible uses of LLM-based technologies can even be lethal---according to a report from the Belgian news outlet La Libre (via Belga News Agency)
~\cite{xiangsuicide2023}, a Belgian man is said to have tragically ended his own life after engaging in conversations with an AI chatbot, for six weeks, discussing climate change. 

Many of the arguments that highlight the potential harms of LLMs are warranted. LLMs are trained on an Internet that is largely devoid of fact-checking. As a result, LLMs often reproduce convincing misinformation~\cite{chen2023can}, and in the context of the COVID-19 pandemic, were found to be capable of generating highly persuasive, difficult to discern health misinformation about COVID-19's precautionary and prevention measures~\cite{zhou2023synthetic}. Some have described LLMs as being similar to super-powerful auto-completion tools, as it can be hard to systematically control for specific outputs~\cite{bender2021dangers}. This can be problematic in a mental health context, where the success of interventions can be highly dependent on the nature of a provider's response~\cite{Glorioso2023, HsuThompson2023, Klepper2023}. LLMs have been described as a ``blurry JPEG image''~\cite{chiang2023} of the rest of the Internet, similarly containing both substantial utility as well as the potential for harm to users.  
Today, just as in digital mental health spaces, there are widespread debates around the consequences of LLMs for truthful public discourse on one hand, and productivity and efficiency on the other.

This article is situated against the backdrop of this larger debate. We present a state-of-the-art informed perspective on contexts where LLMs can potentially be beneficial, and where there may be significant risk of harm. We conclude with directions and recommendations that can help amplify benefits while also minimizing harms.

\section{Theoretical Frameworks}
\begin{figure}[t]
\centering 
\includegraphics[width=.75\linewidth]{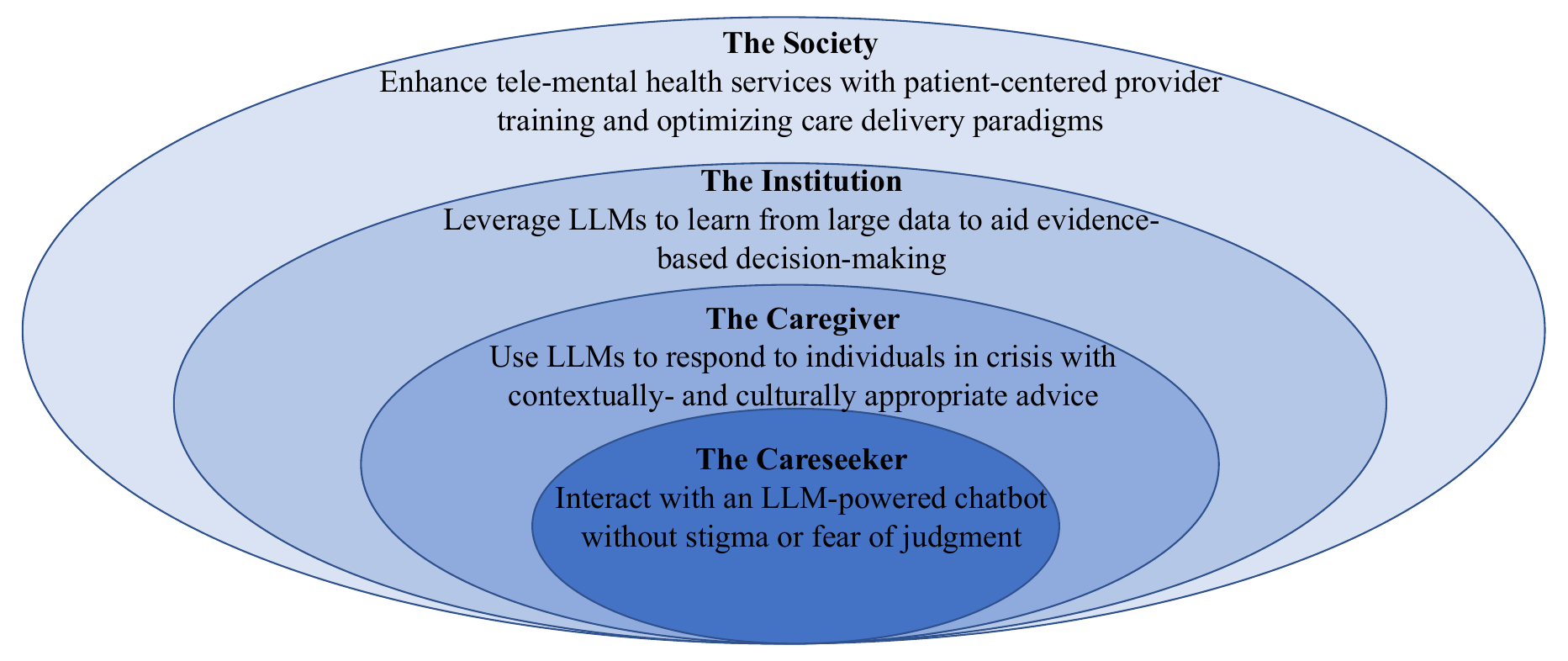}
\caption{An ecological conceptualization of the use of large language models in digital mental health, based on the Social Ecological Model~\cite{reupert2017socio}.}\label{fig:ecological}
\end{figure}

Drawing on Insel's~\cite{insel2023generative} four key areas where LLMs could influence mental health, we conceptualize benefits and harms in the use of LLMs for administering tele-mental healthcare, for supporting crisis response, for providing clinical decision support in psychiatric settings, and for psychotherapy. To organize our discussion, we adopt the Social Ecological Model widely used in the public health field~\cite{reupert2017socio} and map each of Insel's potential application contexts into a four-tier framework. 
This ecological analysis allows us to understand the complex interplay between individuals and the broader social and environmental contexts they are embedded in---contexts that influence and shape people's mental health as well as contexts in which LLM-based technologies could exist. %

Figure~\ref{fig:ecological} shows this organization of the potential uses of LLMs in mental health. The innermost layer focuses on the individual seeking mental healthcare and considers personal factors such as values, beliefs, knowledge, skills, and experiences  in how an individual might appropriate LLMs to seek help and support to overcome their distress. The next layer is the caregiver layer, where LLMs might be appropriated by supporters, counsellors, crisis volunteers, and others to respond to an individual in crisis. The third layer centers around institutional efforts and initiatives in caregiving, where we discuss the role that LLMs can play in enabling decision-making within health systems. At the final layer, we discuss potential futures where LLMs could transform telehealth paradigms. 

In addition, to systematically identify and discuss the factors that can support or hinder caregiving at each level resulting from LLMs, we adopt the concept of affordances~\cite{gibson1977theory}. Affordances refer to the potential actions or uses that an object, environment, or system offers to individuals based on their perceived characteristics or features~\cite{norman1999affordance}. To understand the benefits and harms of LLMs in digital mental health, we begin by describing the various affordances of LLMs that are relevant for the above four-tiered ecological framework. First, LLMs enable \textit{natural language understanding}---users can input text-based queries, prompts, or commands in natural language, and the model interprets and responds to them coherently. Second, by offering end users the ability to engage in open dialogue, and by parsing the semantics of the user input, LLMs allow users to interact with conversational agents in a natural environment~\cite{o2022massive}. These affordances are relevant to the use of LLMs in psychotherapy. Next, LLMs are extremely powerful \textit{information retrieval systems} and are able to parse very large repositories of data efficiently---users can thus request information on a wide range of topics by asking questions or requesting explanations and the model can provide answers, summaries, or context on various subjects. This affordance is relevant to the use of LLMs in crisis response. Third, LLMs can provide \textit{predictions} of outcomes of interest by learning from its underlying data---users can seek answers, recommendations, and suggestions from the LLM for various scenarios and use cases, making this affordance relevant to clinical decision support. Finally, LLMs also allow efficient \textit{text summarization}---users can request concise summaries of longer texts, making it easier to grasp the main points of an article, document, or report. Complementarily, LLMs can be used for content generation and machine translation tasks. Together, these features can be highly valuable for envisioning new models of telehealth. 


\section{For Careseekers: Psychotherapy Chatbot}
Perhaps one of the most widely spoken of applications of LLMs in mental health centers around their use for psychotherapy; Insel discusses the potential for LLMs, particularly GPT-4, to serve as autonomous therapists~\cite{insel2023generative}. In fact, LLM-powered chatbots like Replika are already claiming to be an ``AI comparison, [...] always ready to chat when [a person] needs an empathetic friend''~\cite{newyorkerAITreat}.  
However, for decades, the idea of using a machine to serve as a therapy bot has met with significant heated debate and controversy, specifically around their value to individuals in distress. For example, the Rogerian therapy chatbot ELIZA was first created by Joseph Weizenbaum as a commentary on the irreplaceability of human interaction in mental health support~\cite{weizenbaum1976computer}. After careseekers in distress found ELIZA to provide useful support, some mental health professionals argued that chatbots may be able to scale up basic forms of mental health support~\cite{colby1966computer}. Weizenbaum was shocked by this enthusiasm, and responded that ``\textit{no humane therapy of any kind}'' can or should be done via a chatbot computer program~\cite{weizenbaum1977computers}, grounding his argument in Rogers's own ideas around person-centered therapy. 
These debates continue today, and this section synthesizes the salient points in these conversations.  

\subsection{Potential Benefits}
\paragraph{Improving the Reach of Therapeutic Care.}
Mental health conversational agents have historically been rule-based, meaning they would engage with users based on predetermined scripts~\cite{vaidyam2019chatbots}. This limitation made it challenging for conversational agents to deliver human-like interactions, as they could not engage in open-ended conversations that were tailored to users' emotional requirements. LLMs (grounded in generative AI) have demonstrated impressive performance in participating in realistic human-like conversations in a coherent manner, following practically any type of prompt from the end user~\cite{ma2023understanding}. 

A subsequent benefit offered by this type of naturalistic LLM-powered chatbots in psychotherapy is accessibility. Chatbots are available around the clock, providing users with a convenient and immediate source of support. This can be especially valuable for individuals who have difficulty accessing traditional in-person therapy due to geographical or scheduling constraints~\cite{vallee2022everyday}, those living in underserved areas, such as mental health professional shortage areas~\cite{thomas2009county,rochford2023leveraging}, or those who might be otherwise marginalized in conventional mental health care~\cite{scrutton2017epistemic,pendse2023marginalization}---over half of U.S. counties lack a single psychiatrist~\cite{thomas2009county}, and a recent survey revealed that 60\% of mental health professionals have had no available slots for new patients following the COVID-19 pandemic~\cite{kuehn2022clinician}. Research has demonstrated that online and self-guided single-session interventions can reduce depressive symptoms~\cite{schleider2022randomized, schleider2023little}. The naturalistic and accessible interface associated with LLM-powered chatbots may allow for a new modality to deliver efficacious and self-guided psychotherapy interventions.    

\paragraph{Improving Therapeutic Quality through Personalization and Evidence-based Adaptation.}
LLM-based chatbots are trained on large amounts of historical data, to the order of many terabytes of data~\cite{thirunavukarasu2023large}. Current and future LLMs could be further fine-tuned to be better at providing support based on whether past support responses (from humans or the chatbot) were evaluated by humans as being efficacious. This might include the specific emotional patterns, style, or tonality of response that align best with the needs and expectations of a given client or patient. Researchers have demonstrated that natural language processing techniques that harness the power of LLMs can empower and equip supporters with writing suggestions during practice training sessions. For example, Hsu et al.~\cite{hsu2023helping} create the CARE system for mock chats, which uses LLMs to offer suggestions to online peer supporters as they undergo training. 
Participants found recommendations from the CARE system to be helpful, particularly when faced with uncertainty around how best to respond during the mock training session. Learnings from LLM-based systems can be directly applied when supporters are helping people in distress without the aid of any computational system.
Similarly, Sharma et al.~\cite{sharma2023human} developed HAILEY, an AI-powered agent that offers real-time guidance to enhance the empathetic responses of peer supporters as they assist those in need of support. The research found that peer supporters can effectively incorporate AI feedback with this system, benefiting from both direct and indirect AI assistance, without developing an excessive reliance on AI. In contrast, supporters reported enhanced self-efficacy after receiving AI feedback, underscoring the promise of LLMs as teaching aids for supporters. 
LLMs can further allow for an enhanced ability to use data for actionable and evidence-based insights into the nature of support, including how to adapt support for the individual and their context.    

\paragraph{Destigmatizing the Seeking of Care.}
LLM-based chatbots can be built to draw on evidence-based psychotherapy techniques to deliver helpful support exercises for a user, with potentially less stigma than traditional psychotherapy. Recent empirical research has demonstrated that language models can be utilized to help individuals work to reframe negative thoughts and beliefs through in-context learning~\cite{sharma2023cognitive}, if trained on relevant data. For example, Ziems et al.~\cite{ziems2022inducing} draw upon six theoretically grounded reframing strategies to demonstrate that language models can be used to identify and reframe various types of cognitive distortions. Similarly, 
Sharma et al.~\cite{sharma2023cognitive} utilize a dataset consisting of potential thoughts and corresponding reframed thoughts (validated by practitioners) to train an LLM that generates reframed thoughts for new contexts.\footnote{This system is publicly testable at \href{bit.ly/changing-thoughts}{bit.ly/changing-thoughts}.} This work demonstrates the promise of LLMs in providing CBT-based exercises that are personalized to a user's context, which may lower the barriers to seeking treatment or support, often stemming from stigma, shame, or structural factors, as has been argued for AI as well~\cite{butler2006empirical,palmer2022beneficent}. 
LLM-based chatbots may thus motivate more people to reach out for mental health needs, and help normalize the process of seeking care through self-disclosure and processing of thoughts, whether with other humans or chatbots. Such normalization of care-seeking has been identified to be very important to fight the prevailing societal stigma surrounding mental illness~\cite{wright2009cognitive}. 

\subsection{Potential Harms}
Black~\cite{black2023humanizing} has noted how those working at the intersection of AI and care are quick to explain that ``a chatbot will never equal, let alone surpass, the abilities of a human therapist or counselor.'' However, a growing number of commercial initiatives are building LLM-informed psychotherapy chatbots, with a belief that ``mental health chatbots [can be] instantly and all but universally available at a fraction of the cost of a therapist''~\cite{black2023humanizing}. Given the pace of development, while there may be benefits for those in distress, 
a careful consideration of harms is crucial.

\paragraph{An Erosion of the Therapeutic Alliance.} The therapeutic alliance, characterized by trust, mutual respect, and emotional connection, is a key predictor of psychotherapy outcomes. The emotional connection built between psychotherapists and their clients plays a pivotal role in developing this alliance
~\cite{stanhope2013examining}. Patients are more likely to engage actively in psychotherapy sessions and work towards making behavior changes when they perceive a sense of genuine care and understanding from their therapist~\cite{krupnick2006role}. 

Chatbots may be trained on billions of documents that describe rich emotional experiences, but they lack the capacity for genuine emotional connection. Psychotherapists are trained to not only understand a client's emotional state, but to also empathize with a client's feelings through their own lived emotional experiences. Psychotherapists discern the nuances of complex emotional experiences, including co-occuring sadness, anger, and fear, and respond in a compassionate manner. 
They can also pick up on non-verbal cues -- 
body language, facial expressions, tone of voice, and even pauses in speech are all non-verbal (or textual) cues that psychotherapists interpret~\cite{gladstein1974nonverbal}. LLM powered chatbots, in contrast, are limited in their capacity to understand or interpret the broader context of the client's experiences, as expressed through these non-verbal cues. This limitation makes it more difficult for a connection to be built between chatbot and user. 

The emotional connection between therapist and client is also foundational to building trust and rapport, particularly when clients are assured that their expressions of distress and vulnerability are confidential~\cite{price2017developing, greenberg2006emotion}. This trust is essential for successful psychotherapy, as it encourages open and honest communication. Psychotherapy excels in an environment when patients are comfortable sharing their deepest concerns, fostering a therapeutic alliance~\cite{lambert2001research}. 
To date, it is unclear if chatbots can build the type of trust and rapport that is essential to successful therapy, particularly given the potential for digital mental health data to be leaked, sold, or even legally used for targeted advertising~\cite{kim2023databrokers, ftcbetterhelp2023}. LLM chatbots risk taking away the human element of psychotherapy. 
Existing studies have demonstrated that users express negative reactions to unhelpful and risky chatbot responses~\cite{de2022chatbots}, which is dangerous -- first experiences with mental healthcare can quickly set the tone for future engagements (or disengagements) with mental healthcare~\cite{pendse2021can}.     

\paragraph{A Lack of Appropriate Safeguards for Patient Safety.} The excitement for LLM-based therapy bots needs to be tempered with the reality and safety issues they pose to clients and patients. Generative AI-based conversational agents have been found to be unable to recognize, and respond appropriately to, signs of distress~\cite{de2022chatbots}. Most commercial chatbots for mental health claim psychology-trained workers to carefully write and evaluate the dialogue for these bots~\cite{black2023humanizing}. However, in the interest of safety, chatbots can be constrained by predefined scripts~\cite{jargon2023} and may struggle to adapt to the intricate and evolving emotional states of clients.  Research has also shown that while chatbots can offer structured interventions that can be easily scaled up (such as self-guided exercises from CBT), sustained user engagement often requires the involvement of human psychotherapists~\cite{abd2019overview}. It is also important to note that many mental health issues involve complex and multifaceted emotional experiences. Patients dealing with trauma, grief, or deep-seated emotional struggles often require nuanced, empathetic support that goes beyond providing information or structured interventions~\cite{andersson2014advantages}. Human therapists can adapt and respond to the evolving needs of their clients, maintaining their motivation and commitment to therapy, 
and tailor their responses and interventions to the unique needs of each individual~\cite{norcross2011works}. LLM chatbots may lack capacity for such personalized adaption of therapeutic techniques in response to unique patient needs. 

\section{For Care Providers: Crisis Response}
The introduction of the 988 crisis response number in July 2022 was a significant step forward in the field of mental health~\cite{suran2023new,matthews2023national}. This new service provides individuals in crisis with immediate access to mental health support, as an alternative to calling 911 for police or fire department assistance. 
However, one challenge that remains is the scarcity of a trained workforce to respond effectively to mental health crises, often resulting in calls being transferred to 911 and risking additional harm to callers~\cite{zabelski2023crisis}. Between 2015 to 2020, 23\% of people who were killed by police in the U.S. were experiencing mental health crises at the time~\cite{skorburg_friesen_2021}. Insel~\cite{insel2023generative} highlights the potential of LLMs in addressing this issue, emphasizing their dual role in potentially detecting the severity of a crisis and supporting 988 in providing real-time assistance. Below, we explore the benefits and harms of using LLMs in this context.

\subsection{Potential Benefits}
\paragraph{Matching Users to Contextually Relevant Volunteers.}
Meta-reviews of existing crisis line services have revealed a lack of effectiveness, especially of distal outcomes, such as reduction in symptoms or feelings of distress in callers in a followup period after the call
~\cite{hoffberg2020effectiveness}. Empirical studies of helplines have investigated the reasons driving this phenomenon, finding that they stem from dissatisfied callers and responders inability to attend to callers' diverse needs~\cite{pendse2020like}. 
LLMs could strengthen infrastructures like 988 by helping
to effectively route people in distress to helpline volunteers based on an assessment of their needs.
Through their proficiency in natural language understanding, LLMs could be utilized to analyze the language of distress to 
match people to the types of context-specific support and specialized volunteers that they may need.
This rapid assessment can help ensure that individuals in immediate need receive the contextually relevant forms of support they need promptly. In turn, this could reduce the burden on crisis responders by prioritizing high-risk cases. These possibilities have already been demonstrated in prior research -- Althoff et al.~\cite{althoff2016large} used data from an SMS texting-based counseling service where people in crisis engaged in therapeutic conversations with counselors, to build computational approaches that described which types of language of volunteers elicited better conversational outcomes. 

\paragraph{Culturally-Sensitive Vetted Guidance.} LLM technologies can be harnessed to bolster the expertise of crisis responders and volunteers through real-time linguistic framing and support. This potential is supported by recent research that has trained language models towards semantic, issue-based, and lexical reframing of opinions, arguments, as well as unhelpful thoughts~\cite{chakrabarty2021entrust,maddela2023training}. Additionally, LLMs could help by surfacing guidance and recommendations to specific crisis situations which have been previously vetted (by human experts), to be helpful in mitigating crisis; Sharma and De Choudhury~\cite{sharma2018mental} highlighted this potential through models that learn from positive support seeking and support provisioning engagements on online forums. LLMs could help to suggest appropriate interventions, coping strategies, and de-escalation techniques based on the information provided by the caller and matching this to similar crisis intervention scenarios in historical data. This real-time assistance can be invaluable in calming the situation and ensuring the safety of the individual in crisis. 
Further, we discussed in the previous section that LLMs can be programmed to respond in linguistically diverse ways; in a crisis scenario, culturally resonating support can have significant impact on the caller's mental health outcomes. Prior research has shown how language barriers can hinder effective crisis intervention~\cite{pendse2020like}; thus by empowering crisis volunteers with language tailored to the identity and culture of the caller, LLMs could help promote greater inclusivity on helplines, including 988 as well as those in more resource-constrained settings.

\subsection{Potential Harms}
\paragraph{A Lack of Contextual Understanding.} Although AI-powered crisis response has been advocated to be particularly helpful during rapidly evolving ad well as protracted societal crises like mass shootings~\cite{cheng2020ai} and the COVID-19 pandemic~\cite{abbas2021role} due to their ability to be deployed quickly and at scale, crisis response is an extremely high-stakes domain, and thus risk and harms could have debilitating impacts on stakeholders involved, especially the callers. First, multiple factors influence what precipitates a crisis as well as what strategies could help mitigate it~\cite{murphy2015crisis}. Such factors could exist outside the realm of the training data used to build the LLMs, often perhaps in messy offline contexts---contexts in which LLMs may have little to no insight. Further, the ``black box'' nature of such AIs make identifying contextual gaps inscrutable~\cite{ehsan2022human}. While a human crisis responder could be well-equipped, trained, or use their awareness of the situation to probe those unobserved factors behind the crisis, 
LLMs may provide inappropriate or hallucinated responses or those without sufficient empathy, potentially leading to an ineffective, harmful, or non-consensual crisis response that perhaps even worsens the caller's emotional state. Even with prompt engineering, it can be hard to control what an LLM may say to an individual in crisis---the harmful outputs produced by the AI-assisted NEDA chatbot, as introduced in the Introduction,  demonstrates how harmful directly exposing crisis response service users to LLMs may be. 

\paragraph{The Complexities of Data Use and Consent.} Ultimately, an LLM is only as good as its training data~\cite{bender2021dangers}. Scholars have repeatedly discussed how by learning from large datasets on the internet, LLMs could 
``overrepresent hegemonic viewpoints and encode biases''~\cite{bender2021dangers}, creating ethically contentious outcomes 
potentially extending or even exacerbating inequities in care~\cite{obermeyer2019dissecting}. However, beyond diversity, the scale and scope of the training data also matters, especially in an application domain like crisis response. The successful use of LLMs in crisis responses hinges on being able to train them on copious amounts of data spanning caller-volunteer conversations. Normally, these conversations are seldom recorded or transcribed beyond service optimization purposes, both to protect confidentiality of the data, as well as to ensure callers find the crisis resource to allow more disinhibited exchange with the volunteer~\cite{turkington2020people}. It is known that knowledge of being tracked or monitored could create a ``Hawthorne effect''~\cite{mccambridge2014systematic} leading to people being less truthful of their thoughts and feelings, and perhaps feeling silenced and fearful of the consequences of surveillance. 
Since privacy is often ``contextual''~\cite{nissenbaum2004privacy}, in a crisis setting, callers might be concerned about how their data is collected, stored, and perhaps most importantly, who does what with this data. An emerging crisis may also present challenges to a caller's capacity to recognize these potential harms and to make the most rational decision for themselves. The best example of this might be the Crisis Text Line scandal from 2022, where the efforts of the organization to collect and share conversational data with a for-profit spinoff without user consent alarmed many~\cite{politico2022ctl}. Thus efforts to collect conversational data going forward, to train LLMs, may undermine the goals of adequate assessment of a caller's experience and deploying the most appropriate intervention. Sourcing training data without adequate informed consent or participatory involvement of the data producers (e.g., the people in distress seeking help) may further complicate their use in LLMs, by reducing their agency in controlling ``what data is captured, how it is used, or who it benefits''~\cite{li2023dimensions} and by rendering their data labor invisible~\cite{yoo2024missed}. 

\section{For Institutions: Clinical Decision Support}
In his article, Tom Insel argued that LLMs can provide clinicians with comprehensive and up-to-date information, aiding in the decision-making process~\cite{insel2023generative}. We examine the benefits and potential harms of incorporating LLMs into clinical decision support.

\subsection{Potential Benefits}
\paragraph{Unlocking Vast Medical Knowledge.}
One of the most significant advantages of using LLMs in clinical decision support is the ability to access a wealth of information. These models can learn from vast amounts of medical literature, offering clinicians insights on various conditions, treatments, and potential side effects, including that is latest in the medical field. Side effects of psychiatric medications in particular are often very nuanced and demonstrate patient heterogeneity in effects~\cite{saha2019social, saha2021understanding}. LLMs can not only surface how similar patients have responded to specific treatments but also can help inform health professionals about previously unknown potential side effects by learning from complex drug interactions spanning thousands of clinical trials~\cite{ayvaz2015toward} and online discussions around interactions~\cite{papoutsaki2021understanding}. This knowledge can assist healthcare professionals in making well-informed decisions. 
The knowledge ingested by LLMs can also be utilized toward predictive analytic approaches, in order to augment decision support about patient outcomes, hospital readmission risk, and disease progression---all of which have been shown to be outcome predictable  using machine learning techniques~\cite{lyons1997predicting,alonso2018data,birnbaum2019detecting}. LLMs could both improve the precision of these predictions \textit{and} aid in proactive patient management and resource allocation. 

\paragraph{Providing Individually-Tailored Recommendations.}
LLMs can provide tailored recommendations based on the patient's unique circumstances that is gleaned from their historical electronic health records, clinical notes, or hospital discharge summaries, which together can significantly impact patient care and improve patient outcomes~\cite{yang2022large}. LLMs can easily and quickly ingest diverse types of conventional health (EHR) and health-adjacent data (e.g., smartphone or wearable use, social media activities) of patients to develop such personalized models~\cite{steinberg2021language}, and utilize it for differential diagnosis~\cite{kottlors2023feasibility}. 
When it comes to personalized treatment, differential diagnosis is perhaps one of the biggest strengths offered by LLMs~\cite{kottlors2023feasibility}. 
With this knowledge, clinicians may be empowered to reduce the risk of misdiagnosis~\cite{gala2023utility}; misdiagnosis hurts the efficacy of therapeutic and pharmacologic treatments down the road, and can enable individuals to function better in their personal and professional lives, maintain relationships, and achieve their life goals. 
LLMs can importantly democratize the medical knowledge encoded in interactions amongst health professionals by providing information not only to clinicians but also to patients and their families~\cite{clusmann2023future}. Informed patients can engage in shared decision-making with providers, fostering a collaborative approach to healthcare and improving health literacy~\cite{botelho2023leveraging}.

\subsection{Potential Harms}

\paragraph{Perpetuating Misinformation and Contextually Uninformed Decisions.}
Relying solely on LLMs for clinical decision support without verification from human experts can lead to the dissemination of misinformation, potentially harming patients' health and well-being~\cite{verma2022examining}. Jin and Chandra et al~\cite{jin2023better} recently showed that while GPT-4 like LLMs are largely adept at providing accurate responses to a variety of health queries, for some types of queries they produce incorrect information. In fact, Zhou et al. showed that GPT models (a type of LLM built by Open AI) could be prompt engineered relatively easily to reproduce medically incorrect information~\cite{zhou2023synthetic}. 
Due to the complex and sensitive nature of mental health issues, clinical decision-making demands nuanced, context-specific understanding and personalized care. LLMs, while powerful, lack the ability to grasp the intricacies of an individual's mental state and history, especially factors ans aspects that may not be apparent in its training data such as from EHRs~\cite{bhatt2021universality}. Given the lack of ``objective'' medical measures of mental illness, clinicians utilize a variety of collateral information in their decision-making~\cite{petrik2015balancing}, for instance, through interactions with the patients' family members or relying on non-clinical insights. LLMs are likely to miss opportunities to learn from such collateral information that tend to be heavily individual-specific and unique. By relying on specific types of biased training data stemming from the lived experience of specific (majority) populations, LLMs might overlook the subtleties in language related to mental health, such as expressions tied to traumatic experiences or coping mechanisms, which shape a person's own conceptualization of their mental health~\cite{nichter2010idioms, kleinman1988illness}. As Harrigian et al~\cite{harrigian2020models} noted, when these nuances are not considered during the training of predictive models (here, LLMs), there is a risk of these signals generating numerous false alarms in decision-making when applied to different populations. This may be exacerbated by temporal artifacts as also noted by Harrigian et al~\cite{harrigian2020models}. That is, when there are group-level differences in temporal alignment of the data between model training and deployment, it can exert an impact on predictive performance of LLMs dedicated for psychiatric decision-making.

Moreover, existing commercial LLMs have been demonstrated to not generalize well to non-English health contexts~\cite{jin2023better}, producing to poor quality, non-comprehensive, and hallucinated information~\cite{zhang2023siren}, thus disadvantaging non-English speaking patients. In psychiatry, where accurate diagnosis and timely, culturally-sensitive treatment is paramount for success of care paradigms~\cite{pendse2022treatment}, relying solely on automated systems can be particularly perilous, that can lead to poor response to treatment or prolonged duration of untreated mental illness. 

\paragraph{Suggesting Clinically Unverified or Incorrect Treatments.}
The above problems can be exacerbated when considering individuals with serious mental illnesses, such as schizophrenia or bipolar disorder. 
These conditions often require highly individualized care and diagnosis, as symptoms can manifest differently from person to person. Misinformation from an LLM could lead to inappropriate treatment plans, exacerbating the suffering of already vulnerable individuals. 
LLMs learn from web and social media data, which is often considered a strength, but such data (e.g., Reddit health conversations) has also been shown to include scientifically unsupported, clinically unverified, sometimes dangerous treatments~\cite{chancellor2019opioid, elsherief2021characterizing}. Uncritical use of such data for model training may result in such medically unverified strategies to trickle into an LLM's decision-making~\cite{tate2023chatgpt}. 
Furthermore, 
researchers have shown that in supervised classification tasks, LLMs often fail to outperform existing fine-tuned traditional machine learning models~\cite{ziems2023can}. 
In resource-scare settings where access to mental health professionals and services is limited~\cite{rochford2023leveraging}, these systems may be one of the few available resources for support~\cite{tate2023chatgpt}, making 
false positives or false negatives in clinical decisions make the potential for harm even more significant. False negatives in predicting adverse mental health events may leave marginalized individuals without the support they need. False positives, on the other hand, may exacerbate stigma, perpetuate bias, harassment and marginalization, and importantly, diminish trust in care systems~\cite{balcombe2023ai}. 

\paragraph{Ethical Issues in Automated Decision-Making.}
The use of LLMs for clinical decision-making in mental health also raises ethical and legal questions, particularly in matters of liability~\cite{yoo2021clinician,duffourc2023generative}. If a clinical decision goes awry based on recommendations from an LLM, who bears responsibility? When clinicians rely on LLMs for diagnosis, treatment, or advice without proper verification, they could be held accountable for any resulting harm to patients. Legal questions may emerge regarding their duty of care, professional negligence, and the informed use of technology~\cite{saenz2023autonomous}. Unlike traditional human healthcare providers, LLMs lack the capacity for judgment and accountability~\cite{bender2021dangers,weidinger2021ethical}, which makes determining liability in cases of misinformation or harm resulting from improper decisions challenging. Liability in LLM-based clinical decision-making in mental health---``a field where classifications of diseases as well as definitions of what is and what is not a disease are in a state of constant flux''~\cite{halleck1980psychiatrist} -- may further be complicated by the often unpredictable and context-dependent nature of mental disorders. 
Administrative models have been demonstrated to be accurate in predicting suicidal behaviors at only 50\% rate~\cite{kessler2020suicide}, with scholars often considering this task ``unpredictable''~\cite{large2018role} or suicide events to be ``random''~\cite{soper2022randomness}. Even if LLMs promise better performance in such prediction tasks, as has been the case with other types of AI~\cite{coppersmith2018natural}, liability issues complicate realizing their practical clinical value.

\section{For Society: Telehealth 2.0}
Finally, Insel envisions generative AI 
to revolutionize telehealth services~\cite{insel2023generative} through what he calls ``Telehealth 2.0''. LLMs in telehealth may offer a variety of opportunities and challenges. 

\subsection{Potential Benefits}

\paragraph{Enhancing Efficiency of Care-Delivery.} Perhaps the most promising opportunity pertains to the efficiency and automation provided by LLMs in the context of telehealth. One of the significant benefits of LLMs in mental health can be the automation of various tasks. Since LLMs can be used to transcribe and summarize large volumes of text data, such as stemming from therapy sessions, it can help to reduce the administrative burden on clinicians and healthcare workers~\cite{sezgin2023artificial}. This, in turn, could allow clinicians to focus more on providing quality patient care. Automation may also lead to more consistent and thorough documentation of patient interactions, which can be invaluable in maintaining continuity of care~\cite{khanbhai2022using}. In fact, by identifying patterns, trends, and insights in patient-clinician interactions, the chances of patient concerns being overlooked can be reduced -- an issue widely recognized in traditional healthcare models, and particularly among racial and gender minorities~\cite{hoffman2016racial, chen2008gender, bougie2019influence, lee2019racial}. 
It can instill confidence in the patient that their voices and concerns are more likely to be heard. 

\paragraph{Enriching Counselor/Provider Training with Cultural Competence.} Next, LLMs can play a valuable role in counselor/provider training in the context of telehealth. LLMs can be employed to curate and compile extensive resources, including textbooks, research papers, case studies, and guidelines relevant to counseling. These can serve as a comprehensive knowledge base for trainees~\cite{ravi2023large}. LLMs can also generate realistic case scenarios, reflecting various mental health issues, patient backgrounds, and cultural contexts. Trainees can engage with these scenarios to practice their counseling skills~\cite{safranek2023role}. Notably, LLMs can analyze and provide feedback on the trainee's counseling sessions. As shown in Sharma et al's work~\cite{sharma2021towards, sharma2023human}, by processing the dialogue and the dynamics of the conversation, AI systems can offer insights on communication effectiveness, active listening, and therapeutic rapport. This feedback can be invaluable for trainees to identify areas for improvement. Moreover, such a scalable approach can ensure that a broader group of individuals is equipped to respond effectively to those in need. Furthermore, as discussed by Pendse et al~\cite{pendse2022treatment}, one of the biggest impediments to accessing care lies in a poor alignment between a patient's cultural understandings of their experience of mental health, and that of the counsellor/provider's. LLMs can assist in cultural competency training by providing trainees with information about various cultures, their values, and belief systems. 

\subsection{Potential Harms}

\paragraph{Dehumanizing Mental Health Treatment Paradigms.} Automation by learning from vast datasets has been touted to be a inimitable strength of LLMs~\cite{li2023chatgpt}. However, while automation can be efficient, it may also lead to a depersonalized healthcare experience and by rendering human labor (of healthcare workers) increasingly obsolescent. 
Mental health care is not only about diagnosing and treating the underlying illnesses but also about providing emotional support, empathy, and comfort to patients~\cite{gateshill2011attitudes}. These human aspects of care in an LLM-informed telehealth model may not be fully replicated by machines~\cite{montemayor2022principle}. Moreover, the success of many pharmacological and therapeutic approaches to mental health treatment hinges upon trust and respect between patients and providers~\cite{brown2009trust}. A dehumanized healthcare experience punctuated by AI may erode this trust, as patients might feel that their care is being delivered by algorithms that lack a comprehensive understanding of their individual needs. Careless automation of telehealth paradigms may also lead to disinvestment in care work and to a shift away from patients' values and preferences --  interfering with the goals of a patient-centered model of care~\cite{carey2016beyond}. By removing humans from the care delivery loop, increased LLM or generative AI based automation may dissuade individuals from pursuing the healthcare profession altogether or demotivate healthcare workers from persisting in a profession where their jobs may be at risk of displacement. The obsolescence of healthcare workers could have  economic and social implications, including job losses, economic dislocation, and potential labor market disruptions.  


\paragraph{Threats to Data Privacy.} The most prominent risk perhaps centers around that the analysis of sensitive patient data by LLMs -- an approach that can threaten privacy~\cite{weidinger2021ethical}. This is especially critical if the utility of LLMs is viewed to be centered around gleaning meaning from electronic health records, which contain patient health information (PHI)~\cite{wornow2023shaky}. It has been noted in recent research that LLMs can accidentally divulge private information if prompted in specific ways~\cite{chen2023can}. This is because, in learning from vast data, these models learn relationships between personally identifiable information or PII (e.g., names, addresses etc.) and other linguistic elements. These attributes of LLMs can make them vulnerable to data breaches~\cite{kasneci2023chatgpt} to unauthorized parties or to bad actors employing them for contexts beyond the intended use. 

\paragraph{Demographic and Representational Biases.} Language models have, for some years, been shown to replicate human-like as well as systemic biases, whether around gender~\cite{bartl-etal-2020-unmasking, bhardwaj2021investigating, kurita-etal-2019-measuring} 
or racial sterotypes~\cite{tan2019assessing, caliskan2017semantics, kiritchenko-mohammad-2018-examining}. 
In a telehealth context, if used to provide recommendations, such as to clinicians or trainees based on historical patient-clinician interactions or therapy sessions, LLMs can inadvertently introduce biases based on any underlying skewness in the training data. If not carefully monitored, this can lead to unfair or inaccurate representations of patient-provider interactions. 
For instance, if the training data skews the representativeness of one demographic group versus another, LLM-based suggestions in telehealth could lead to or exacerbate gender, racial, or ethnic inequities. This concern is not merely theoretical; commercially accessible LLMs have exhibited racial and gender biases in non-medical contexts, and these very models have been found to propagate stereotypes related to race within the field of medicine~\cite{omiye2023beyond}. Biases may also arise from LLMs learning stigmatizing representations of language in training data~\cite{tate2023generative}, and propagating those through inappropriate language or portraying mental health issues in a negative light. It is already known that stigmatizing language often surfaces in EHR patient notes~\cite{weiner2023incidence}. If used for knowledge summarization in a telehealth context, LLMs may thus over-pathologize common emotions or behaviors, causing undue alarm. Stigma already hinders support-seeking in mental health, and this is known to prolong the duration of untreated mental illness~\cite{mueser2020clinical}. 
Additional concerns of harm stem from LLMs creating informational or perspective ``echo chambers'', where 
LLMs inadvertently perpetuate pre-existing beliefs and biases held by providers or patients, as they exist in biased training data. For example, it is already known that screening tools in psychiatry may be biased in ways that tend to over-diagnose Black patients with schizophrenia~\cite{barnes2013race}. LLMs may pathologize these biases by learning from such data. 
An over-reliance on LLMs for summarizing EHR information or providing training materials could result in not challenging end users sufficiently to consider a variety of approaches to providing mental healthcare, rather than an LLM-recommended approach. 

\section{Conclusion, Recommendations, and Future Directions}

Long before LLMs were put in the hands of the lay internet user through ChatGPT, Bender and Koller~\cite{bender2020climbing} noted that it will be crucial to acknowledge the limitations of LLMs and place their strengths within a broader perspective. This approach can serve to moderate exaggerated expectations, which can lead both the general public and researchers astray in terms of the capabilities of these technologies. 
At the same time, these understandings have the potential to stimulate fresh research pathways that are not solely reliant on the utilization of ever bigger language models in every possible domain of societal interest. 
Throughout this article, we discussed the many dimensions of the debate centered around the use of LLMs in digital mental health applications. We offer some reflections and specific considerations for future research.

\subsection{Reflections and Lessons Learned}
What is apparent from our above discussion is that, however fine-tuned and tailored LLMs may be to data stemming from real-world mental health contexts, LLM-powered chatbots or decision-support tools cannot serve as a replacement for human psychotherapists or health workers. Neither can a machine alone be a surrogate to a real person during moments of distress. 
The significance of the human therapist may be further underscored by the fact that the appeal of chatbots in therapy may vary among different age groups. For digital natives, who are more accustomed to interacting with technology, the appeal of a machine therapist might be greater that of previous generations; but across age groups, preferences for a human therapist is likely to remain strong.
Thus, what are the boundaries of the role of LLMs in digital mental health, for who, and what responsibilities do developers and healthcare providers have in ensuring their ethical use? A realistic ``safety-first'' approach might be to use them as surrogates, rather than as standalone AI therapists. 

For such a safety-first approach, it will be essential to strike a balance between harnessing the potential of LLMs and ensuring that human experts remain integral to the decision-making process in psychiatry and mental health, particularly when dealing with the most vulnerable and resource-scarce populations. An approach, such as a human-in-the-loop or more preferably, an AI-in-the-loop system, may help to combine the cognitive strengths of healthcare providers with the analytical capabilities of LLMs. Horvitz's~\cite{horvitz1999principles} conceptualization of ``mixed initiative'' systems might be particularly pertinent to mental health applications where, based on the situation, users can take the lead when they have a specific goal or intention, while also allowing AI to take the initiative when it can provide value or assist the user. In critical uses surrounding clinical decision-support or crisis response, mixed initiatives between the human and the AI (LLM) can enable them to work together as partners, with the system actively seeking input from the user and the user having the ability to request information, clarification, or assistance from the LLM component. 
To this end, it will be essential to establish clear protocols for verifying the underlying LLM's recommendations of clinical decisions or crisis intervention strategies, ensuring human experts remain responsible for final decisions and that providers maintain their ethical and legal obligations to prioritize patient safety and well-being. In psychotherapy or telehealth, human oversight will  ensure that there are appropriate safeguards in place that prepare for potential harms when the underlying ``stochastic parrots''~\cite{bender2021dangers}, provide inappropriate, incorrect, or misinformed outcomes, because of the non-deterministic nature of these technologies.

In addition, if LLMs are to be utilized in this high stakes domain, thoughtful investments to create or gather realistic
training data will also be needed, that do not compromise the very mission and values that underlie psychotherapeutic practices or crisis intervention. For instance, creating keystone datasets has been advocated to help advance psychological research using LLMs~\cite{demszky2023using}. Consent, awareness, and literacy regarding how specific data (e.g., EHRs, crisis helpline call logs, or psychotherapy chat transcripts) is used in specific LLMs, how, and by whom will be equally important considerations in ensuring governance of these systems. On that note, for any type of mental health-relevant data that can be made available for LLM training, utmost care will need to be employed to prevent unauthorized access or data breaches, which could lead to severe harm, including identity theft or emotional distress for individuals in distress or in crisis. Within the telehealth context, in employing LLMs to summarize patient-provider interactions or therapy sessions to support clinical care or to train counsellors, it will be paramount to protect PHI and PII in EHR data, ensuring not only compliance with data protection regulations, but also to maintain patient trust their personal information is secure and will not be misused.




Broadly speaking, there is a pressing requirement for increased empirical research and advocacy efforts aimed at helping mental health service users and practitioners differentiate the quality, usability, and efficacy of LLMs, as well as identifying the suitable use cases, scenarios, and target populations that stand to gain from their application. 

\subsection{Recommendations}
Building on these reflections, we suggest technical, ethical, and human-centered recommendations to ensure that the use of LLMs in mental health settings carefully balances effectiveness and responsibility.

Developers of LLM-based mental health tools should shoulder the responsibility in keeping applications safe for end users of these tools. This can be implemented through self-accountability frameworks, such as, by reporting to the public, or through outcomes of red teaming efforts. Additional accountability can come from companies describing what types of training data was used to build the models, how they were evaluated following training, what performance metrics were used to assess performance, and the extent to which performance was tests in various contexts and scenarios. Several efforts in these lines have been proposed in the algorithmic fairness and accountability literature, such as Datasheets for describing the capabilities and limits of datasets used for building AI models~\cite{gebru2021datasheets}, Model Cards for bringing transparency to how complex models work~\cite{mitchell2019model}, and disclosures of ethics practices to demonstrate how model builders remain accountable for the outcomes of AI~\cite{ajmani2023systematic}.   
Some stakeholders have argued that this can be achieved if the creators of domain-specific LLMs open source their models~\cite{digidayCaseAgainst}, others advocating for a regulatory solution~\cite{forbesCouncilPost}. And yet, some have critiqued both approaches because of the risk of open LLMs being exploited by malicious actors or regulation resulting in concentrating power in the hands of a few that stifles innovation and competition~\cite{mozillaJointStatement}. Nevertheless, what this debate underscores is a need for continuous monitoring and evaluation mechanisms to ensure responsible usage and adherence to ethical guidelines within a high stakes domain like mental health.

There has been some policy movement around developing standards and tools to help ensure that AI technologies are safe, secure, and trustworthy; e.g., US President Joe Biden's 2023 executive order on AI~\cite{biden2023executive} or the Group of Seven (G7) announcement of a new code of conduct and international guiding principles on AI~\cite{europaPressCorner}. However, scholars have argued that a one-size-fits-all regulatory model for generative AI will be unsuitable for specific health applications~\cite{mesko2023imperative}, and an adaptable approach to oversight will be needed that can evolve with the rapidly and ever-evolving capabilities of this technology. This adaptive approach will need to go hand in hand with continuous monitoring, external auditing, and benchmarked evaluation of these systems to ensure responsible usage and adherence to ethical guidelines.

Autonomous bodies to enforce oversight would also need to be created in the digital mental health field to establish what types of standards might be suitable for the four types of uses of LLMs in mental health. For instance, what standards would ensure safety if LLMs were to be integrated into the 988 system or within psychotherapy contexts? Pharmacological treatments undergo a clearance process with the FDA and scholars have long advocated for a need to consider similar regulatory processes for digital mental health as well. After many years of research, Empatica is one of the few digital mental health technologies that has received clearance from the FDA for medical use~\cite{empatica2023, empatica2022}. 
This conversation may be extended to LLM-based mental health technologies too, to ensure that a reasonable level of safety is promised in any application that reaches mental health support seekers. 

Furthermore, it will be essential for legal, infrastructure, privacy, and security teams to review organizational policies and procedures to guarantee adherence to state and federal laws and regulations, particularly in the context of personal health information exchange protocols, accountability, liability, service reimbursement, and clinical workflows. Concurrently, there is a demand for the creation of educational curricula and methods to instruct people with lived experience of mental illness, mental health professionals and caregivers, and health system administrators in fundamentals of generative AI, its practical application, and its role in enhancing how care is extended.

In the book \textit{The Soul of Care}~\cite{kleinman2020soul}, Arthur Kleinman describes how ``the work of the doctor has moved away from hands-on practice to high-technology diagnosis and treatment,'' which has distanced doctors from engaging with the human experiences of their patients. Medical enterprises, digital healthcare services, and healthcare institutions have initiated the integration of LLMs into their core operations~\cite{mesko2023imperative}. In this piece, we emphasize the need to attend to immediate challenges in the use of this paradigm-shifting technology, given its rapid clinical roll-out~\cite{mesko2023imperative}. As Arthur Kleinman discusses, the importance of emotionally engaged human doctors in the practice of care can never be replaced, and future LLM-augmented medical technologies must be cognizant of the importance of human connection in care. 


\section*{Acknowledgments and Disclosures}

This work was supported by NIMH grants R01MH117172 (PI: De Choudhury) and P50MH115838 (Co-I: De Choudhury), and a grant from the American Foundation for Suicide Prevention (PI: De Choudhury). This content is solely the responsibility of the authors and does not necessarily represent the official views of the National Institutes of Mental Health or AFSP. We thank the members of the Social Dynamics and Well-Being Lab at Georgia Tech for helping to shape early discussions on the topic.

\printbibliography

@article{openai2022chatgpt,
    author    = {OpenAI},
    title     = {Introducing ChatGPT},
    month     = {November},
    year      = {2022},
    url       = {https://openai.com/blog/chatgpt}
}

@article{bhardwaj2021investigating,
  title={Investigating gender bias in bert},
  author={Bhardwaj, Rishabh and Majumder, Navonil and Poria, Soujanya},
  journal={Cognitive Computation},
  volume={13},
  number={4},
  pages={1008--1018},
  year={2021},
  publisher={Springer}
}

@article{tan2019assessing,
  title={Assessing social and intersectional biases in contextualized word representations},
  author={Tan, Yi Chern and Celis, L Elisa},
  journal={Advances in neural information processing systems},
  volume={32},
  year={2019}
}

@misc{empatica2022,
  author = {{Empatica}},
  title = {Empatica Receives New FDA Clearance for Its Health Monitoring Platform and Announces Series B Financing},
  year = {2022},
  month = {November},
  url = {https://www.prnewswire.com/news-releases/empatica-receives-new-fda-clearance-for-its-health-monitoring-platform-and-announces-series-b-financing-301685344.html},
}

@book{schleider2023little,
  title={Little Treatments, Big Effects: How to Build Meaningful Moments that Can Transform Your Mental Health},
  author={Schleider, Jessica},
  year={2023},
  publisher={Robinson}
}

@article{schleider2022randomized,
  title={A randomized trial of online single-session interventions for adolescent depression during COVID-19},
  author={Schleider, Jessica L and Mullarkey, Michael C and Fox, Kathryn R and Dobias, Mallory L and Shroff, Akash and Hart, Erica A and Roulston, Chantelle A},
  journal={Nature Human Behaviour},
  volume={6},
  number={2},
  pages={258--268},
  year={2022},
  publisher={Nature Publishing Group UK London}
}

@misc{empatica2023,
  author = {{Empatica}},
  title = {Empatica's Platform Receives New FDA Clearance for Cardiac Digital Biomarkers},
  month = {November},
  year = {2023},
  url = {https://www.prnewswire.com/news-releases/empaticas-platform-receives-new-fda-clearance-for-cardiac-digital-biomarkers-301975974.html},
}

@inproceedings{kiritchenko-mohammad-2018-examining,
    title = "Examining Gender and Race Bias in Two Hundred Sentiment Analysis Systems",
    author = "Kiritchenko, Svetlana  and
      Mohammad, Saif",
    editor = "Nissim, Malvina  and
      Berant, Jonathan  and
      Lenci, Alessandro",
    booktitle = "Proceedings of the Seventh Joint Conference on Lexical and Computational Semantics",
    month = jun,
    year = "2018",
    address = "New Orleans, Louisiana",
    publisher = "Association for Computational Linguistics",
    url = "https://aclanthology.org/S18-2005",
    doi = "10.18653/v1/S18-2005",
    pages = "43--53",
}

@article{caliskan2017semantics,
  title={Semantics derived automatically from language corpora contain human-like biases},
  author={Caliskan, Aylin and Bryson, Joanna J and Narayanan, Arvind},
  journal={Science},
  volume={356},
  number={6334},
  pages={183--186},
  year={2017},
  publisher={American Association for the Advancement of Science}
}

@inproceedings{kurita-etal-2019-measuring,
    title = "Measuring Bias in Contextualized Word Representations",
    author = "Kurita, Keita  and
      Vyas, Nidhi  and
      Pareek, Ayush  and
      Black, Alan W  and
      Tsvetkov, Yulia",
    editor = "Costa-juss{\`a}, Marta R.  and
      Hardmeier, Christian  and
      Radford, Will  and
      Webster, Kellie",
    booktitle = "Proceedings of the First Workshop on Gender Bias in Natural Language Processing",
    month = aug,
    year = "2019",
    address = "Florence, Italy",
    publisher = "Association for Computational Linguistics",
    url = "https://aclanthology.org/W19-3823",
    doi = "10.18653/v1/W19-3823",
    pages = "166--172",
}

@inproceedings{bartl-etal-2020-unmasking,
    title = "Unmasking Contextual Stereotypes: Measuring and Mitigating {BERT}{'}s Gender Bias",
    author = "Bartl, Marion  and
      Nissim, Malvina  and
      Gatt, Albert",
    editor = "Costa-juss{\`a}, Marta R.  and
      Hardmeier, Christian  and
      Radford, Will  and
      Webster, Kellie",
    booktitle = "Proceedings of the Second Workshop on Gender Bias in Natural Language Processing",
    month = dec,
    year = "2020",
    address = "Barcelona, Spain (Online)",
    publisher = "Association for Computational Linguistics",
    url = "https://aclanthology.org/2020.gebnlp-1.1",
    pages = "1--16",
}

@article{koutsouleris2022promise,
  title={From promise to practice: towards the realisation of AI-informed mental health care},
  author={Koutsouleris, Nikolaos and Hauser, Tobias U and Skvortsova, Vasilisa and De Choudhury, Munmun},
  journal={The Lancet Digital Health},
  volume={4},
  number={11},
  pages={e829--e840},
  year={2022},
  publisher={Elsevier}
}

@article{omiye2023beyond,
  title={Beyond the hype: large language models propagate race-based medicine},
  author={Omiye, Jesutofunmi A and Lester, Jenna and Spichak, Simon and Rotemberg, Veronica and Daneshjou, Roxana},
  journal={medRxiv},
  pages={2023--07},
  year={2023},
  publisher={Cold Spring Harbor Laboratory Press}
}

@article{tate2023generative,
  title={Generative Artificial Intelligence Tools in Medicine Will Amplify Stigmatizing Language},
  author={Tate, Steven},
  journal={Journal of Addiction Medicine},
  pages={10--1097},
  year={2023},
  publisher={LWW}
}

@article{de2022chatbots,
  title={Chatbots and Mental Health: Insights into the Safety of Generative AI},
  author={De Freitas, Julian and U{\u{g}}uralp, Ahmet Kaan and O{\u{g}}uz-U{\u{g}}uralp, Zeliha and Puntoni, Stefano},
  journal={Journal of Consumer Psychology},
  year={2022},
  publisher={Wiley Online Library}
}

@inproceedings{pendse2021can,
  title={“Can I not be suicidal on a Sunday?”: understanding technology-mediated pathways to mental health support},
  author={Pendse, Sachin R and Sharma, Amit and Vashistha, Aditya and De Choudhury, Munmun and Kumar, Neha},
  booktitle={Proceedings of the 2021 CHI Conference on Human Factors in Computing Systems},
  pages={1--16},
  year={2021}
}

@phdthesis{black2023humanizing,
  title={De-humanizing Care: An Ethnography of Mental Health Artificial Intelligence},
  author={Black, Valerie E},
  year={2023},
  school={University of California, Berkeley}
}

@article{krupnick2006role,
  title={The role of the therapeutic alliance in psychotherapy and pharmacotherapy outcome: Findings in the National Institute of Mental Health Treatment of Depression Collaborative Research Program},
  author={Krupnick, Janice L and Sotsky, Stuart M and Elkin, Irene and Simmens, Sam and Moyer, Janet and Watkins, John and Pilkonis, Paul A},
  journal={Focus},
  volume={64},
  number={2},
  pages={532--277},
  year={2006},
  publisher={Am Psychiatric Assoc}
}

@article{stanhope2013examining,
  title={Examining the relationship between choice, therapeutic alliance and outcomes in mental health services},
  author={Stanhope, Victoria and Barrenger, Stacey L and Salzer, Mark S and Marcus, Stephen C},
  journal={Journal of Personalized Medicine},
  volume={3},
  number={3},
  pages={191--202},
  year={2013},
  publisher={MDPI}
}

@article{gladstein1974nonverbal,
  title={Nonverbal communication and counseling/psychotherapy: A review},
  author={Gladstein, Gerald A},
  journal={The Counseling Psychologist},
  volume={4},
  number={3},
  pages={34--57},
  year={1974},
  publisher={Sage Publications Sage CA: Thousand Oaks, CA}
}

@article{price2017developing,
  title={Developing patient rapport, trust and therapeutic relationships},
  author={Price, Bob},
  journal={Nursing Standard},
  volume={31},
  number={50},
  year={2017},
  publisher={RCN Publishing Company Limited}
}

@article{greenberg2006emotion,
  title={Emotion in psychotherapy: A practice-friendly research review},
  author={Greenberg, Leslie S and Pascual-Leone, Antonio},
  journal={Journal of clinical psychology},
  volume={62},
  number={5},
  pages={611--630},
  year={2006},
  publisher={Wiley Online Library}
}

@article{lambert2001research,
  title={Research summary on the therapeutic relationship and psychotherapy outcome.},
  author={Lambert, Michael J and Barley, Dean E},
  journal={Psychotherapy: Theory, research, practice, training},
  volume={38},
  number={4},
  pages={357},
  year={2001},
  publisher={Division of Psychotherapy (29), American Psychological Association}
}

@article{abd2019overview,
  title={An overview of the features of chatbots in mental health: A scoping review},
  author={Abd-Alrazaq, Alaa A and Alajlani, Mohannad and Alalwan, Ali Abdallah and Bewick, Bridgette M and Gardner, Peter and Househ, Mowafa},
  journal={International Journal of Medical Informatics},
  volume={132},
  pages={103978},
  year={2019},
  publisher={Elsevier}
}

@article{andersson2014advantages,
  title={Advantages and limitations of Internet-based interventions for common mental disorders},
  author={Andersson, Gerhard and Titov, Nickolai},
  journal={World Psychiatry},
  volume={13},
  number={1},
  pages={4--11},
  year={2014},
  publisher={Wiley Online Library}
}

@article{norcross2011works,
  title={What works for whom: Tailoring psychotherapy to the person},
  author={Norcross, John C and Wampold, Bruce E},
  journal={Journal of clinical psychology},
  volume={67},
  number={2},
  pages={127--132},
  year={2011},
  publisher={Wiley Online Library}
}

@article{vaidyam2019chatbots,
  title={Chatbots and conversational agents in mental health: a review of the psychiatric landscape},
  author={Vaidyam, Aditya Nrusimha and Wisniewski, Hannah and Halamka, John David and Kashavan, Matcheri S and Torous, John Blake},
  journal={The Canadian Journal of Psychiatry},
  volume={64},
  number={7},
  pages={456--464},
  year={2019},
  publisher={Sage Publications Sage CA: Los Angeles, CA}
}

@article{ma2023understanding,
  title={Understanding the benefits and challenges of using large language model-based conversational agents for mental well-being support},
  author={Ma, Zilin and Mei, Yiyang and Su, Zhaoyuan},
  journal={arXiv preprint arXiv:2307.15810},
  year={2023}
}

@article{o2022massive,
  title={Massive data language models and conversational artificial intelligence: Emerging issues},
  author={O’Leary, Daniel E},
  journal={Intelligent Systems in Accounting, Finance and Management},
  volume={29},
  number={3},
  pages={182--198},
  year={2022},
  publisher={Wiley Online Library}
}

@article{vallee2022everyday,
  title={Everyday geography and service accessibility: the contours of disadvantage in relation to mental health},
  author={Vall{\'e}e, Julie and Shareck, Martine and Kestens, Yan and Frohlich, Katherine L},
  journal={Annals of the American Association of Geographers},
  volume={112},
  number={4},
  pages={931--947},
  year={2022},
  publisher={Taylor \& Francis}
}

@article{thomas2009county,
  title={County-level estimates of mental health professional shortage in the United States},
  author={Thomas, Kathleen C and Ellis, Alan R and Konrad, Thomas R and Holzer, Charles E and Morrissey, Joseph P},
  journal={Psychiatric services},
  volume={60},
  number={10},
  pages={1323--1328},
  year={2009},
  publisher={Am Psychiatric Assoc}
}

@article{rochford2023leveraging,
  title={Leveraging symptom search data to understand disparities in US mental health care: demographic analysis of search engine trace data},
  author={Rochford, Ben and Pendse, Sachin and Kumar, Neha and De Choudhury, Munmun},
  journal={JMIR Mental Health},
  volume={10},
  pages={e43253},
  year={2023},
  publisher={JMIR Publications Toronto, Canada}
}

@article{pendse2023marginalization,
  title={Marginalization and the Construction of Mental Illness Narratives Online: Foregrounding Institutions in Technology-Mediated Care},
  author={Pendse, Sachin R and Kumar, Neha and De Choudhury, Munmun},
  journal={Proceedings of the ACM on Human-Computer Interaction},
  volume={7},
  number={CSCW2},
  pages={1--30},
  year={2023},
  publisher={ACM New York, NY, USA}
}

@incollection{scrutton2017epistemic,
  title={Epistemic injustice and mental illness},
  author={Scrutton, Anastasia Philippa},
  booktitle={The Routledge handbook of epistemic injustice},
  pages={347--355},
  year={2017},
  publisher={Routledge}
}

@article{kuehn2022clinician,
  title={Clinician shortage exacerbates pandemic-fueled “mental health crisis”},
  author={Kuehn, Bridget M},
  journal={JAMA},
  volume={327},
  number={22},
  pages={2179--2181},
  year={2022},
  publisher={American Medical Association}
}

@article{thieme2020machine,
  title={Machine learning in mental health: A systematic review of the HCI literature to support the development of effective and implementable ML systems},
  author={Thieme, Anja and Belgrave, Danielle and Doherty, Gavin},
  journal={ACM Transactions on Computer-Human Interaction (TOCHI)},
  volume={27},
  number={5},
  pages={1--53},
  year={2020},
  publisher={ACM New York, NY, USA}
}

@article{gibson1977theory,
  title={The theory of affordances},
  author={Gibson, James J},
  journal={Hilldale, USA},
  volume={1},
  number={2},
  pages={67--82},
  year={1977}
}

@article{norman1999affordance,
  title={Affordance, conventions, and design},
  author={Norman, Donald A},
  journal={interactions},
  volume={6},
  number={3},
  pages={38--43},
  year={1999},
  publisher={ACM New York, NY, USA}
}

@article{chakrabarty2021entrust,
  title={ENTRUST: Argument reframing with language models and entailment},
  author={Chakrabarty, Tuhin and Hidey, Christopher and Muresan, Smaranda},
  journal={arXiv preprint arXiv:2103.06758},
  year={2021}
}

@article{maddela2023training,
  title={Training Models to Generate, Recognize, and Reframe Unhelpful Thoughts},
  author={Maddela, Mounica and Ung, Megan and Xu, Jing and Madotto, Andrea and Foran, Heather and Boureau, Y-Lan},
  journal={arXiv preprint arXiv:2307.02768},
  year={2023}
}

@article{althoff2016large,
  title={Large-scale analysis of counseling conversations: An application of natural language processing to mental health},
  author={Althoff, Tim and Clark, Kevin and Leskovec, Jure},
  journal={Transactions of the Association for Computational Linguistics},
  volume={4},
  pages={463--476},
  year={2016},
  publisher={MIT Press One Rogers Street, Cambridge, MA 02142-1209, USA journals-info~…}
}

@article{palmer2022beneficent,
  title={Beneficent dehumanization: Employing artificial intelligence and carebots to mitigate shame-induced barriers to medical care},
  author={Palmer, Amitabha and Schwan, David},
  journal={Bioethics},
  volume={36},
  number={2},
  pages={187--193},
  year={2022},
  publisher={Wiley Online Library}
}

@article{suran2023new,
  title={How the new 988 lifeline is helping millions in mental health crisis},
  author={Suran, Melissa},
  journal={JAMA},
  year={2023}
}

@article{matthews2023national,
  title={National preparedness for 988—the new mental health emergency hotline in the United States},
  author={Matthews, Samantha and Cantor, Jonathan H and Holliday, Stephanie Brooks and Bialas, Armenda and Eberhart, Nicole K and Breslau, Joshua and McBain, Ryan K},
  journal={Preventive medicine reports},
  volume={33},
  pages={102208},
  year={2023},
  publisher={Elsevier}
}

@article{lyons1997predicting,
  title={Predicting psychiatric emergency admissions and hospital outcome},
  author={Lyons, John S and Stutesman, John and Neme, Janice and Vessey, John T and O'Mahoney, Michael T and Camper, H Joseph},
  journal={Medical care},
  pages={792--800},
  year={1997},
  publisher={JSTOR}
}

@article{alonso2018data,
  title={Data mining algorithms and techniques in mental health: a systematic review},
  author={Alonso, Susel G{\'o}ngora and de La Torre-D{\'\i}ez, Isabel and Hamrioui, Sofiane and L{\'o}pez-Coronado, Miguel and Barreno, Diego Calvo and Nozaleda, Lola Mor{\'o}n and Franco, Manuel},
  journal={Journal of medical systems},
  volume={42},
  pages={1--15},
  year={2018},
  publisher={Springer}
}

@article{hoffberg2020effectiveness,
  title={The effectiveness of crisis line services: a systematic review},
  author={Hoffberg, Adam S and Stearns-Yoder, Kelly A and Brenner, Lisa A},
  journal={Frontiers in public health},
  volume={7},
  pages={399},
  year={2020},
  publisher={Frontiers Media SA}
}

@article{abbas2021role,
  title={The role of social media in the advent of COVID-19 pandemic: crisis management, mental health challenges and implications},
  author={Abbas, Jaffar and Wang, Dake and Su, Zhaohui and Ziapour, Arash},
  journal={Risk management and healthcare policy},
  pages={1917--1932},
  year={2021},
  publisher={Taylor \& Francis}
}

@article{cheng2020ai,
  title={AI-Powered mental health chatbots: Examining users’ motivations, active communicative action and engagement after mass-shooting disasters},
  author={Cheng, Yang and Jiang, Hua},
  journal={Journal of Contingencies and Crisis Management},
  volume={28},
  number={3},
  pages={339--354},
  year={2020},
  publisher={Wiley Online Library}
}

@inproceedings{ehsan2022human,
  title={Human-Centered Explainable AI (HCXAI): beyond opening the black-box of AI},
  author={Ehsan, Upol and Wintersberger, Philipp and Liao, Q Vera and Watkins, Elizabeth Anne and Manger, Carina and Daum{\'e} III, Hal and Riener, Andreas and Riedl, Mark O},
  booktitle={CHI conference on human factors in computing systems extended abstracts},
  pages={1--7},
  year={2022}
}

@article{murphy2015crisis,
  title={Crisis intervention for people with severe mental illnesses},
  author={Murphy, Suzanne M and Irving, Claire B and Adams, Clive E and Waqar, Muhammad},
  journal={Cochrane Database of Systematic Reviews},
  number={12},
  year={2015},
  publisher={John Wiley \& Sons, Ltd}
}

@article{obermeyer2019dissecting,
  title={Dissecting racial bias in an algorithm used to manage the health of populations},
  author={Obermeyer, Ziad and Powers, Brian and Vogeli, Christine and Mullainathan, Sendhil},
  journal={Science},
  volume={366},
  number={6464},
  pages={447--453},
  year={2019},
  publisher={American Association for the Advancement of Science}
}

@article{skorburg_friesen_2021,
  title = {Mind the Gaps: Ethical and Epistemic Issues in the Digital Mental Health Response to Covid-19},
  author = {Skorburg, Joshua August and Friesen, Phoebe},
  journal = {Hastings Center Report},
  volume = {51},
  number = {6},
  pages = {23},
  year = {2021}
}

@book{kleinman2020soul,
  title={The soul of care: the moral education of a husband and a doctor},
  author={Kleinman, Arthur},
  year={2020},
  publisher={Penguin}
}

@article{turkington2020people,
  title={Why do people call crisis helplines? Identifying taxonomies of presenting reasons and discovering associations between these reasons},
  author={Turkington, Robin and Mulvenna, Maurice D and Bond, Raymond R and O’Neill, Siobhan and Potts, Courtney and Armour, Cherie and Ennis, Edel and Millman, Catherine},
  journal={Health informatics journal},
  volume={26},
  number={4},
  pages={2597--2613},
  year={2020},
  publisher={SAGE Publications Sage UK: London, England}
}

@article{yang2022large,
  title={A large language model for electronic health records},
  author={Yang, Xi and Chen, Aokun and PourNejatian, Nima and Shin, Hoo Chang and Smith, Kaleb E and Parisien, Christopher and Compas, Colin and Martin, Cheryl and Costa, Anthony B and Flores, Mona G and others},
  journal={NPJ Digital Medicine},
  volume={5},
  number={1},
  pages={194},
  year={2022},
  publisher={Nature Publishing Group UK London}
}

@article{steinberg2021language,
  title={Language models are an effective representation learning technique for electronic health record data},
  author={Steinberg, Ethan and Jung, Ken and Fries, Jason A and Corbin, Conor K and Pfohl, Stephen R and Shah, Nigam H},
  journal={Journal of biomedical informatics},
  volume={113},
  pages={103637},
  year={2021},
  publisher={Elsevier}
}

@article{kottlors2023feasibility,
  title={Feasibility of differential diagnosis based on imaging patterns using a large language model},
  author={Kottlors, Jonathan and Bratke, Grischa and Rauen, Philip and Kabbasch, Christoph and Persigehl, Thorsten and Schlamann, Marc and Lennartz, Simon},
  journal={Radiology},
  volume={308},
  number={1},
  pages={e231167},
  year={2023},
  publisher={Radiological Society of North America}
}

@article{gala2023utility,
  title={The utility of language models in cardiology: a narrative review of the benefits and concerns of ChatGPT-4},
  author={Gala, Dhir and Makaryus, Amgad N},
  journal={International Journal of Environmental Research and Public Health},
  volume={20},
  number={15},
  pages={6438},
  year={2023},
  publisher={MDPI}
}

@article{clusmann2023future,
  title={The future landscape of large language models in medicine},
  author={Clusmann, Jan and Kolbinger, Fiona R and Muti, Hannah Sophie and Carrero, Zunamys I and Eckardt, Jan-Niklas and Laleh, Narmin Ghaffari and L{\"o}ffler, Chiara Maria Lavinia and Schwarzkopf, Sophie-Caroline and Unger, Michaela and Veldhuizen, Gregory P and others},
  journal={Communications Medicine},
  volume={3},
  number={1},
  pages={141},
  year={2023},
  publisher={Nature Publishing Group UK London}
}

@article{bhatt2021universality,
  title={On the universality of deep contextual language models},
  author={Bhatt, Shaily and Goyal, Poonam and Dandapat, Sandipan and Choudhury, Monojit and Sitaram, Sunayana},
  journal={arXiv preprint arXiv:2109.07140},
  year={2021}
}

@article{duffourc2023generative,
  title={Generative AI in health care and liability risks for physicians and safety concerns for patients},
  author={Duffourc, Mindy and Gerke, Sara},
  journal={Jama},
  year={2023}
}

@article{saenz2023autonomous,
  title={Autonomous AI systems in the face of liability, regulations and costs},
  author={Saenz, Agustina D and Harned, Zach and Banerjee, Oishi and Abr{\`a}moff, Michael D and Rajpurkar, Pranav},
  journal={NPJ digital medicine},
  volume={6},
  number={1},
  pages={185},
  year={2023},
  publisher={Nature Publishing Group UK London}
}

@article{weidinger2021ethical,
  title={Ethical and social risks of harm from language models},
  author={Weidinger, Laura and Mellor, John and Rauh, Maribeth and Griffin, Conor and Uesato, Jonathan and Huang, Po-Sen and Cheng, Myra and Glaese, Mia and Balle, Borja and Kasirzadeh, Atoosa and others},
  journal={arXiv preprint arXiv:2112.04359},
  year={2021}
}

@article{kessler2020suicide,
  title={Suicide prediction models: a critical review of recent research with recommendations for the way forward},
  author={Kessler, Ronald C and Bossarte, Robert M and Luedtke, Alex and Zaslavsky, Alan M and Zubizarreta, Jose R},
  journal={Molecular psychiatry},
  volume={25},
  number={1},
  pages={168--179},
  year={2020},
  publisher={Nature Publishing Group UK London}
}

@article{large2018role,
  title={The role of prediction in suicide prevention},
  author={Large, Matthew Michael},
  journal={Dialogues in clinical neuroscience},
  year={2018},
  publisher={Taylor \& Francis}
}

@article{halleck1980psychiatrist,
  title={The Psychiatrist’s Liability for Negligent Diagnosis},
  author={Halleck, Seymour L and Halleck, Seymour L},
  journal={Law in the Practice of Psychiatry: A Handbook for Clinicians},
  pages={65--82},
  year={1980},
  publisher={Springer}
}

@article{soper2022randomness,
  title={On the randomness of suicide: An evolutionary, clinical call to transcend suicide risk assessment},
  author={Soper, CA and Malo Ocejo, P and Large, Matthew M},
  journal={Evolutionary psychiatry: evolutionary perspectives on mental health},
  pages={134--152},
  year={2022},
  publisher={Cambridge University Press and Royal College of Psychiatrists}
}

@article{coppersmith2018natural,
  title={Natural language processing of social media as screening for suicide risk},
  author={Coppersmith, Glen and Leary, Ryan and Crutchley, Patrick and Fine, Alex},
  journal={Biomedical informatics insights},
  volume={10},
  pages={1178222618792860},
  year={2018},
  publisher={SAGE Publications Sage UK: London, England}
}

@article{yoo2021clinician,
  title={Clinician perspectives on using computational mental health insights from patients’ social media activities: design and qualitative evaluation of a prototype},
  author={Yoo, Dong Whi and Ernala, Sindhu Kiranmai and Saket, Bahador and Weir, Domino and Arenare, Elizabeth and Ali, Asra F and Van Meter, Anna R and Birnbaum, Michael L and Abowd, Gregory D and De Choudhury, Munmun},
  journal={JMIR Mental Health},
  volume={8},
  number={11},
  pages={e25455},
  year={2021},
  publisher={JMIR Publications Toronto, Canada}
}

@article{sezgin2023artificial,
  title={Artificial intelligence in healthcare: Complementing, not replacing, doctors and healthcare providers},
  author={Sezgin, Emre},
  journal={Digital Health},
  volume={9},
  pages={20552076231186520},
  year={2023},
  publisher={SAGE Publications Sage UK: London, England}
}

@article{khanbhai2022using,
  title={Using natural language processing to understand, facilitate and maintain continuity in patient experience across transitions of care},
  author={Khanbhai, Mustafa and Warren, Leigh and Symons, Joshua and Flott, Kelsey and Harrison-White, Stephanie and Manton, Dave and Darzi, Ara and Mayer, Erik},
  journal={International journal of medical informatics},
  volume={157},
  pages={104642},
  year={2022},
  publisher={Elsevier}
}

@article{ravi2023large,
  title={Large Language Models and Medical Education: Preparing for a Rapid Transformation in How Trainees Will Learn to Be Doctors},
  author={Ravi, Akshay and Neinstein, Aaron and Murray, Sara G},
  journal={ATS Scholar},
  pages={ats--scholar},
  year={2023},
  publisher={American Thoracic Society}
}

@misc{safranek2023role,
  title={The role of large language models in medical education: applications and implications},
  author={Safranek, Conrad W and Sidamon-Eristoff, Anne Elizabeth and Gilson, Aidan and Chartash, David},
  journal={JMIR Medical Education},
  volume={9},
  pages={e50945},
  year={2023},
  publisher={JMIR Publications Toronto, Canada}
}

@article{wornow2023shaky,
  title={The shaky foundations of large language models and foundation models for electronic health records},
  author={Wornow, Michael and Xu, Yizhe and Thapa, Rahul and Patel, Birju and Steinberg, Ethan and Fleming, Scott and Pfeffer, Michael A and Fries, Jason and Shah, Nigam H},
  journal={npj Digital Medicine},
  volume={6},
  number={1},
  pages={135},
  year={2023},
  publisher={Nature Publishing Group UK London}
}

@inproceedings{bender2020climbing,
  title={Climbing towards NLU: On meaning, form, and understanding in the age of data},
  author={Bender, Emily M and Koller, Alexander},
  booktitle={Proceedings of the 58th annual meeting of the association for computational linguistics},
  pages={5185--5198},
  year={2020}
}

@article{ziems2023can,
  title={Can Large Language Models Transform Computational Social Science?},
  author={Ziems, Caleb and Held, William and Shaikh, Omar and Chen, Jiaao and Zhang, Zhehao and Yang, Diyi},
  journal={arXiv preprint arXiv:2305.03514},
  year={2023}
}

@inproceedings{balcombe2023ai,
  title={AI Chatbots in Digital Mental Health},
  author={Balcombe, Luke},
  booktitle={Informatics},
  volume={10},
  number={4},
  pages={82},
  year={2023},
  organization={MDPI}
}

@inproceedings{horvitz1999principles,
  title={Principles of mixed-initiative user interfaces},
  author={Horvitz, Eric},
  booktitle={Proceedings of the SIGCHI conference on Human Factors in Computing Systems},
  pages={159--166},
  year={1999}
}

@misc{mozillaJointStatement,
	author = {},
	title = {{J}oint {S}tatement on {A}{I} {S}afety and {O}penness --- open.mozilla.org},
	howpublished = {\url{https://open.mozilla.org/letter/}},
	year = {},
	note = {[Accessed 05-11-2023]},
}

@misc{digidayCaseAgainst,
	author = {Sara Guaglione},
	title = {{T}he case for and against open-source large language models for use in newsrooms --- digiday.com},
	howpublished = {\url{https://digiday.com/media/the-case-for-and-against-open-source-large-language-models-for-use-in-newsrooms/}},
	year = {},
	note = {[Accessed 05-11-2023]},
}

@misc{forbesCouncilPost,
	author = {Stefan Harrer},
	title = {{C}ouncil {P}ost: {F}rom {B}oring {A}nd {S}afe {T}o {E}xciting {A}nd {D}angerous: {W}hy {L}arge {L}anguage {M}odels {N}eed {T}o {B}e {R}egulated --- forbes.com},
	howpublished = {\url{https://www.forbes.com/sites/forbestechcouncil/2023/03/22/from-boring-and-safe-to-exciting-and-dangerous-why-large-language-models-need-to-be-regulated/}},
	year = {},
	note = {[Accessed 05-11-2023]},
}

@article{choi2020development,
  title={Development of a machine learning model using multiple, heterogeneous data sources to estimate weekly US suicide fatalities},
  author={Choi, Daejin and Sumner, Steven A and Holland, Kristin M and Draper, John and Murphy, Sean and Bowen, Daniel A and Zwald, Marissa and Wang, Jing and Law, Royal and Taylor, Jordan and others},
  journal={JAMA network open},
  volume={3},
  number={12},
  pages={e2030932--e2030932},
  year={2020},
  publisher={American Medical Association}
}

@article{elsherief2021characterizing,
  title={Characterizing and identifying the prevalence of web-based misinformation relating to medication for opioid use disorder: Machine learning approach},
  author={ElSherief, Mai and Sumner, Steven A and Jones, Christopher M and Law, Royal K and Kacha-Ochana, Akadia and Shieber, Lyna and Cordier, LeShaundra and Holton, Kelly and De Choudhury, Munmun},
  journal={Journal of medical Internet research},
  volume={23},
  number={12},
  pages={e30753},
  year={2021},
  publisher={JMIR Publications Toronto, Canada}
}

@article{ayers2023evaluating,
  title={Evaluating Artificial Intelligence Responses to Public Health Questions},
  author={Ayers, John W and Zhu, Zechariah and Poliak, Adam and Leas, Eric C and Dredze, Mark and Hogarth, Michael and Smith, Davey M},
  journal={JAMA Network Open},
  volume={6},
  number={6},
  pages={e2317517--e2317517},
  year={2023},
  publisher={American Medical Association}
}

@article{verma2022examining,
  title={Examining the impact of sharing COVID-19 misinformation online on mental health},
  author={Verma, Gaurav and Bhardwaj, Ankur and Aledavood, Talayeh and De Choudhury, Munmun and Kumar, Srijan},
  journal={Scientific Reports},
  volume={12},
  number={1},
  pages={8045},
  year={2022},
  publisher={Nature Publishing Group UK London}
}

@article{zhang2023siren,
  title={Siren's Song in the AI Ocean: A Survey on Hallucination in Large Language Models},
  author={Zhang, Yue and Li, Yafu and Cui, Leyang and Cai, Deng and Liu, Lemao and Fu, Tingchen and Huang, Xinting and Zhao, Enbo and Zhang, Yu and Chen, Yulong and others},
  journal={arXiv preprint arXiv:2309.01219},
  year={2023}
}

@inproceedings{ajmani2023systematic,
  title={A Systematic Review of Ethics Disclosures in Predictive Mental Health Research},
  author={Ajmani, Leah Hope and Chancellor, Stevie and Mehta, Bijal and Fiesler, Casey and Zimmer, Michael and De Choudhury, Munmun},
  booktitle={Proceedings of the 2023 ACM Conference on Fairness, Accountability, and Transparency},
  pages={1311--1323},
  year={2023}
}

@article{gebru2021datasheets,
  title={Datasheets for datasets},
  author={Gebru, Timnit and Morgenstern, Jamie and Vecchione, Briana and Vaughan, Jennifer Wortman and Wallach, Hanna and Iii, Hal Daum{\'e} and Crawford, Kate},
  journal={Communications of the ACM},
  volume={64},
  number={12},
  pages={86--92},
  year={2021},
  publisher={ACM New York, NY, USA}
}

@inproceedings{mitchell2019model,
  title={Model cards for model reporting},
  author={Mitchell, Margaret and Wu, Simone and Zaldivar, Andrew and Barnes, Parker and Vasserman, Lucy and Hutchinson, Ben and Spitzer, Elena and Raji, Inioluwa Deborah and Gebru, Timnit},
  booktitle={Proceedings of the conference on fairness, accountability, and transparency},
  pages={220--229},
  year={2019}
}

@misc{newyorkerAITreat,
	author = {Dhruv Khullar},
	title = {{C}an {A}.{I}. {T}reat {M}ental {I}llness? --- newyorker.com},
	howpublished = {\url{https://www.newyorker.com/magazine/2023/03/06/can-ai-treat-mental-illness}},
	year = {},
	note = {[Accessed 06-11-2023]},
}

@misc{europaPressCorner,
	author = {},
	title = {{P}ress corner --- ec.europa.eu},
	howpublished = {\url{https://ec.europa.eu/commission/presscorner/detail/en/ip_23_5379}},
	year = {},
	note = {[Accessed 06-11-2023]},
}

@article{demszky2023using,
  title={Using large language models in psychology},
  author={Demszky, Dorottya and Yang, Diyi and Yeager, David S and Bryan, Christopher J and Clapper, Margarett and Chandhok, Susannah and Eichstaedt, Johannes C and Hecht, Cameron and Jamieson, Jeremy and Johnson, Meghann and others},
  journal={Nature Reviews Psychology},
  pages={1--14},
  year={2023},
  publisher={Nature Publishing Group US New York}
}

@article{barnes2013race,
  title={Race and schizophrenia diagnoses in four types of hospitals},
  author={Barnes, Arnold},
  journal={Journal of Black Studies},
  volume={44},
  number={6},
  pages={665--681},
  year={2013},
  publisher={Sage Publications Sage CA: Los Angeles, CA}
}

@article{mesko2023imperative,
  title={The imperative for regulatory oversight of large language models (or generative AI) in healthcare},
  author={Mesk{\'o}, Bertalan and Topol, Eric J},
  journal={npj Digital Medicine},
  volume={6},
  number={1},
  pages={120},
  year={2023},
  publisher={Nature Publishing Group UK London}
}

@article{kasneci2023chatgpt,
  title={ChatGPT for good? On opportunities and challenges of large language models for education},
  author={Kasneci, Enkelejda and Se{\ss}ler, Kathrin and K{\"u}chemann, Stefan and Bannert, Maria and Dementieva, Daryna and Fischer, Frank and Gasser, Urs and Groh, Georg and G{\"u}nnemann, Stephan and H{\"u}llermeier, Eyke and others},
  journal={Learning and individual differences},
  volume={103},
  pages={102274},
  year={2023},
  publisher={Elsevier}
}

@article{botelho2023leveraging,
  title={Leveraging ChatGPT to Democratize and Decolonize Global Surgery: Large Language Models for Small Healthcare Budgets},
  author={Botelho, Fabio and Tshimula, Jean Marie and Poenaru, Dan},
  journal={World Journal of Surgery},
  pages={1--2},
  year={2023},
  publisher={Springer}
}

@article{birnbaum2019detecting,
  title={Detecting relapse in youth with psychotic disorders utilizing patient-generated and patient-contributed digital data from facebook},
  author={Birnbaum, Michael Leo and Ernala, Sindhu Kiranmai and Rizvi, AF and Arenare, Elizabeth and R. Van Meter, A and De Choudhury, M and Kane, John M},
  journal={NPJ schizophrenia},
  volume={5},
  number={1},
  pages={17},
  year={2019},
  publisher={Nature Publishing Group UK London}
}

@article{zabelski2023crisis,
  title={Crisis lines: current status and recommendations for research and policy},
  author={Zabelski, Sasha and Kaniuka, Andr{\'e}a R and A. Robertson, Ryan and Cramer, Robert J},
  journal={Psychiatric services},
  volume={74},
  number={5},
  pages={505--512},
  year={2023},
  publisher={Am Psychiatric Assoc}
}

@inproceedings{zhou2023synthetic,
  title={Synthetic lies: Understanding ai-generated misinformation and evaluating algorithmic and human solutions},
  author={Zhou, Jiawei and Zhang, Yixuan and Luo, Qianni and Parker, Andrea G and De Choudhury, Munmun},
  booktitle={Proceedings of the 2023 CHI Conference on Human Factors in Computing Systems},
  pages={1--20},
  year={2023}
}

@article{thirunavukarasu2023large,
  title={Large language models in medicine},
  author={Thirunavukarasu, Arun James and Ting, Darren Shu Jeng and Elangovan, Kabilan and Gutierrez, Laura and Tan, Ting Fang and Ting, Daniel Shu Wei},
  journal={Nature medicine},
  volume={29},
  number={8},
  pages={1930--1940},
  year={2023},
  publisher={Nature Publishing Group US New York}
}

@article{hsu2023helping,
  title={Helping the Helper: Supporting Peer Counselors via AI-Empowered Practice and Feedback},
  author={Hsu, Shang-Ling and Shah, Raj Sanjay and Senthil, Prathik and Ashktorab, Zahra and Dugan, Casey and Geyer, Werner and Yang, Diyi},
  journal={arXiv preprint arXiv:2305.08982},
  year={2023}
}

@article{kim2023databrokers,
  title={Data Brokers and the Sale of Americans’ Mental Health Data: The Exchange of Our Most Sensitive Data and What It Means for Personal Privacy},
  author={Kim, Joanne},
  journal={Duke Sanford Cyber Policy Program},
  month={February},
  year={2023}
}

@article{sharma2023cognitive,
  title={Cognitive Reframing of Negative Thoughts through Human-Language Model Interaction},
  author={Sharma, Ashish and Rushton, Kevin and Lin, Inna Wanyin and Wadden, David and Lucas, Khendra G and Miner, Adam S and Nguyen, Theresa and Althoff, Tim},
  journal={arXiv preprint arXiv:2305.02466},
  year={2023}
}

@article{ziems2022inducing,
  title={Inducing positive perspectives with text reframing},
  author={Ziems, Caleb and Li, Minzhi and Zhang, Anthony and Yang, Diyi},
  journal={arXiv preprint arXiv:2204.02952},
  year={2022}
}

@article{butler2006empirical,
  title={The empirical status of cognitive-behavioral therapy: A review of meta-analyses},
  author={Butler, Andrew C and Chapman, Jason E and Forman, Evan M and Beck, Aaron T},
  journal={Clinical psychology review},
  volume={26},
  number={1},
  pages={17--31},
  year={2006},
  publisher={Elsevier}
}

@book{wright2009cognitive,
  title={Cognitive-behavior therapy for severe mental illness: An illustrated guide},
  author={Wright, Jesse H},
  year={2009},
  publisher={American Psychiatric Pub}
}

@misc{tate2023chatgpt,
  title={The ChatGPT therapist will see you now: Navigating generative artificial intelligence's potential in addiction medicine research and patient care},
  author={Tate, Steven and Fouladvand, Sajjad and Chen, Jonathan H and Chen, Chwen-Yuen Angie},
  journal={Addiction},
  year={2023},
  publisher={Wiley Online Library}
}

@article{mueser2020clinical,
  title={Clinical and demographic correlates of stigma in first-episode psychosis: The impact of duration of untreated psychosis},
  author={Mueser, Kim T and DeTore, Nicole R and Kredlow, M Alexandra and Bourgeois, Michelle L and Penn, David L and Hintz, Kathryn},
  journal={Acta Psychiatrica Scandinavica},
  volume={141},
  number={2},
  pages={157--166},
  year={2020},
  publisher={Wiley Online Library}
}

@article{li2023chatgpt,
  title={ChatGPT in Healthcare: A Taxonomy and Systematic Review},
  author={Li, Jianning and Dada, Amin and Kleesiek, Jens and Egger, Jan},
  journal={medRxiv},
  pages={2023--03},
  year={2023},
  publisher={Cold Spring Harbor Laboratory Press}
}

@article{gateshill2011attitudes,
  title={Attitudes towards mental disorders and emotional empathy in mental health and other healthcare professionals},
  author={Gateshill, Georgina and Kucharska-Pietura, Kate and Wattis, John},
  journal={The Psychiatrist},
  volume={35},
  number={3},
  pages={101--105},
  year={2011},
  publisher={Cambridge University Press}
}

@article{weiner2023incidence,
  title={The incidence and disparities in use of stigmatizing language in clinical notes for patients with substance use disorder},
  author={Weiner, Scott G and Lo, Ying-Chih and Carroll, Aleta D and Zhou, Li and Ngo, Ashley and Hathaway, David B and Rodriguez, Claudia P and Wakeman, Sarah E},
  journal={Journal of addiction medicine},
  pages={10--1097},
  year={2023},
  publisher={LWW}
}

@article{montemayor2022principle,
  title={In principle obstacles for empathic AI: why we can’t replace human empathy in healthcare},
  author={Montemayor, Carlos and Halpern, Jodi and Fairweather, Abrol},
  journal={AI \& society},
  volume={37},
  number={4},
  pages={1353--1359},
  year={2022},
  publisher={Springer}
}

@article{brown2009trust,
  title={Trust in mental health services: A neglected concept},
  author={Brown, Patrick and Calnan, Michael and Scrivener, Amanda and Szmukler, George},
  journal={Journal of Mental Health},
  volume={18},
  number={5},
  pages={449--458},
  year={2009},
  publisher={Taylor \& Francis}
}

@article{carey2016beyond,
  title={Beyond patient-centered care: enhancing the patient experience in mental health services through patient-perspective care},
  author={Carey Prof, Timothy A},
  journal={Patient Experience Journal},
  volume={3},
  number={2},
  pages={46--49},
  year={2016}
}

@article{sharma2023human,
  title={Human--AI collaboration enables more empathic conversations in text-based peer-to-peer mental health support},
  author={Sharma, Ashish and Lin, Inna W and Miner, Adam S and Atkins, David C and Althoff, Tim},
  journal={Nature Machine Intelligence},
  volume={5},
  number={1},
  pages={46--57},
  year={2023},
  publisher={Nature Publishing Group UK London}
}

@misc{reupert2017socio,
  title={A socio-ecological framework for mental health and well-being},
  author={Reupert, Andrea},
  journal={Advances in Mental Health},
  volume={15},
  number={2},
  pages={105--107},
  year={2017},
  publisher={Taylor \& Francis}
}

@article{hauser2022promise,
  title={The promise of a model-based psychiatry: building computational models of mental ill health},
  author={Hauser, Tobias U and Skvortsova, Vasilisa and De Choudhury, Munmun and Koutsouleris, Nikolaos},
  journal={The Lancet Digital Health},
  volume={4},
  number={11},
  pages={e816--e828},
  year={2022},
  publisher={Elsevier}
}

@article{radford2018improving,
  title={Improving language understanding by generative pre-training},
  author={Radford, Alec and Narasimhan, Karthik and Salimans, Tim and Sutskever, Ilya},
  year={2018},
  publisher={OpenAI}
}

@article{politico2022ctl, 
  title={Crisis Text Line ends data-sharing relationship with for-profit spinoff}, 
  author={Hendel, John}, 
  year={2022}, 
  month={January},
  url={https://www.politico.com/news/2022/01/31/crisis-text-line-ends-data-sharing-00004001},
  publisher={Politico}
}

@article{papoutsaki2021understanding,
  title={Understanding Delivery of Collectively Built Protocols in an Online Health Community for Discontinuation of Psychiatric Drugs},
  author={Papoutsaki, Alexandra and So, Samuel and Kenderova, Georgia and Shapiro, Bryan and Epstein, Daniel A},
  journal={Proceedings of the ACM on Human-Computer Interaction},
  volume={5},
  number={CSCW2},
  pages={1--29},
  year={2021},
  publisher={ACM New York, NY, USA}
}

@article{ayvaz2015toward,
  title={Toward a complete dataset of drug--drug interaction information from publicly available sources},
  author={Ayvaz, Serkan and Horn, John and Hassanzadeh, Oktie and Zhu, Qian and Stan, Johann and Tatonetti, Nicholas P and Vilar, Santiago and Brochhausen, Mathias and Samwald, Matthias and Rastegar-Mojarad, Majid and others},
  journal={Journal of biomedical informatics},
  volume={55},
  pages={206--217},
  year={2015},
  publisher={Elsevier}
}

@book{kleinman1988illness,
  title={The illness narratives: Suffering, healing, and the human condition},
  author={Kleinman, Arthur},
  year={1988},
  publisher={Basic books}
}

@article{nichter2010idioms,
  title={Idioms of distress revisited},
  author={Nichter, Mark},
  journal={Culture, Medicine, and Psychiatry},
  volume={34},
  pages={401--416},
  year={2010},
  publisher={Springer}
}

@article{petrik2015balancing,
  title={Balancing patient care and confidentiality: considerations in obtaining collateral information},
  author={Petrik, Megan L and Billera, Melodi and Kaplan, Yuliya and Matarazzo, Bridget and Wortzel, Hal},
  journal={Journal of Psychiatric Practice{\textregistered}},
  volume={21},
  number={3},
  pages={220--224},
  year={2015},
  publisher={LWW}
}

@misc{jin2023better,
      title={Better to Ask in English: Cross-Lingual Evaluation of Large Language Models for Healthcare Queries}, 
      author={Yiqiao Jin and Mohit Chandra and Gaurav Verma and Yibo Hu and Munmun De Choudhury and Srijan Kumar},
      year={2023},
      eprint={2310.13132},
      archivePrefix={arXiv},
      primaryClass={cs.CL}
}

@article{ftcbetterhelp2023,
  title={FTC to Ban BetterHelp from Revealing Consumers’ Data, Including Sensitive Mental Health Information, to Facebook and Others for Targeted Advertising},
  author={{Federal Trade Commission}},
  journal={Press Release, Federal Trade Commission},
  month={March},
  day={2},
  year={2023}
}

@article{saha2021understanding,
  title={Understanding side effects of antidepressants: Large-scale longitudinal study on social media data},
  author={Saha, Koustuv and Torous, John and Kiciman, Emre and De Choudhury, Munmun and others},
  journal={JMIR mental health},
  volume={8},
  number={3},
  pages={e26589},
  year={2021},
  publisher={JMIR Publications Inc., Toronto, Canada}
}

@inproceedings{saha2019social,
  title={A social media study on the effects of psychiatric medication use},
  author={Saha, Koustuv and Sugar, Benjamin and Torous, John and Abrahao, Bruno and K{\i}c{\i}man, Emre and De Choudhury, Munmun},
  booktitle={Proceedings of the International AAAI Conference on Web and Social Media},
  volume={13},
  pages={440--451},
  year={2019}
}

@article{nissenbaum2004privacy,
  title={Privacy as contextual integrity},
  author={Nissenbaum, Helen},
  journal={Wash. L. Rev.},
  volume={79},
  pages={119},
  year={2004},
  publisher={HeinOnline}
}

@article{mccambridge2014systematic,
  title={Systematic review of the Hawthorne effect: new concepts are needed to study research participation effects},
  author={McCambridge, Jim and Witton, John and Elbourne, Diana R},
  journal={Journal of clinical epidemiology},
  volume={67},
  number={3},
  pages={267--277},
  year={2014},
  publisher={Elsevier}
}

@inproceedings{pendse2020like,
  title={" Like Shock Absorbers": understanding the human infrastructures of technology-mediated mental health support},
  author={Pendse, Sachin R and Lalani, Faisal M and De Choudhury, Munmun and Sharma, Amit and Kumar, Neha},
  booktitle={Proceedings of the 2020 CHI Conference on Human Factors in Computing Systems},
  pages={1--14},
  year={2020}
}

@article{ziegler2019fine,
  title={Fine-tuning language models from human preferences},
  author={Ziegler, Daniel M and Stiennon, Nisan and Wu, Jeffrey and Brown, Tom B and Radford, Alec and Amodei, Dario and Christiano, Paul and Irving, Geoffrey},
  journal={arXiv preprint arXiv:1909.08593},
  year={2019}
}

@article{weizenbaum1976computer,
  title={Computer power and human reason: From judgment to calculation.},
  author={Weizenbaum, Joseph},
  year={1976},
  publisher={WH Freeman \& Co}
}

@article{weizenbaum1977computers,
  title={Computers as" Therapists"},
  author={Weizenbaum, Joseph},
  journal={Science},
  volume={198},
  number={4315},
  pages={354--354},
  year={1977},
  publisher={American Association for the Advancement of Science}
}

@inproceedings{sharma2021towards,
  title={Towards facilitating empathic conversations in online mental health support: A reinforcement learning approach},
  author={Sharma, Ashish and Lin, Inna W and Miner, Adam S and Atkins, David C and Althoff, Tim},
  booktitle={Proceedings of the Web Conference 2021},
  pages={194--205},
  year={2021}
}

@article{chen2023can,
  title={Can Language Models be Instructed to Protect Personal Information?},
  author={Chen, Yang and Mendes, Ethan and Das, Sauvik and Xu, Wei and Ritter, Alan},
  journal={arXiv preprint arXiv:2310.02224},
  year={2023}
}

@article{chen2008gender,
  title={Gender disparity in analgesic treatment of emergency department patients with acute abdominal pain},
  author={Chen, Esther H and Shofer, Frances S and Dean, Anthony J and Hollander, Judd E and Baxt, William G and Robey, Jennifer L and Sease, Keara L and Mills, Angela M},
  journal={Academic Emergency Medicine},
  volume={15},
  number={5},
  pages={414--418},
  year={2008},
  publisher={Wiley Online Library}
}

@article{bougie2019influence,
  title={Influence of race/ethnicity on prevalence and presentation of endometriosis: a systematic review and meta-analysis},
  author={Bougie, Olga and Yap, Ma I and Sikora, Lindsey and Flaxman, Teresa and Singh, Sukhbir},
  journal={BJOG: An International Journal of Obstetrics \& Gynaecology},
  volume={126},
  number={9},
  pages={1104--1115},
  year={2019},
  publisher={Wiley Online Library}
}

@article{lee2019racial,
  title={Racial and ethnic disparities in the management of acute pain in US emergency departments: meta-analysis and systematic review},
  author={Lee, Paulyne and Le Saux, Maxine and Siegel, Rebecca and Goyal, Monika and Chen, Chen and Ma, Yan and Meltzer, Andrew C},
  journal={The American journal of emergency medicine},
  volume={37},
  number={9},
  pages={1770--1777},
  year={2019},
  publisher={Elsevier}
}

@article{hoffman2016racial,
  title={Racial bias in pain assessment and treatment recommendations, and false beliefs about biological differences between blacks and whites},
  author={Hoffman, Kelly M and Trawalter, Sophie and Axt, Jordan R and Oliver, M Norman},
  journal={Proceedings of the National Academy of Sciences},
  volume={113},
  number={16},
  pages={4296--4301},
  year={2016},
  publisher={National Acad Sciences}
}

@article{reardon2023chatbots,
  title     = {AI Chatbots Could Help Provide Therapy, but Caution Is Needed},
  author    = {Sara Reardon},
  journal   = {Scientific American},
  year      = {2023},
  month     = {June},
  day       = {14},
  url       = {https://www.scientificamerican.com/article/ai-chatbots-could-help-provide-therapy-but-caution-is-needed/},
  note      = {Accessed: 2023-11-03}
}

@misc{biden2023executive,
  title        = {Executive Order on the Safe, Secure, and Trustworthy Development and Use of Artificial Intelligence},
  author       = {Joe Biden},
  year         = {2023},
  howpublished = {The White House},
  url          = {https://www.whitehouse.gov/briefing-room/presidential-actions/2023/10/30/executive-order-on-the-safe-secure-and-trustworthy-development-and-use-of-artificial-intelligence/},
  note         = {Accessed: 2023-11-03}
}

@inproceedings{harrigian2020models,
  title={Do models of mental health based on social media data generalize?},
  author={Harrigian, Keith and Aguirre, Carlos and Dredze, Mark},
  booktitle={Findings of the association for computational linguistics: EMNLP 2020},
  pages={3774--3788},
  year={2020}
}

@incollection{insel2023generative,
    title = {Generative AI and Mental Health},
    author = {Insel, Tom},
    booktitle = {AI Anthology},
    editor = {Horvitz, Eric},
    url = {https://unlocked.microsoft.com/ai-anthology/tom-insel},
    year = {2023},
    month = {June 26}
}

@article{colby1966computer,
  title={A computer method of psychotherapy: Preliminary communication},
  author={Colby, Kenneth Mark and Watt, James B and Gilbert, John P},
  journal={The Journal of Nervous and Mental Disease},
  volume={142},
  number={2},
  pages={148--152},
  year={1966},
  publisher={LWW}
}

@inproceedings{pendse2022treatment,
  title={From treatment to healing: envisioning a decolonial digital mental health},
  author={Pendse, Sachin R and Nkemelu, Daniel and Bidwell, Nicola J and Jadhav, Sushrut and Pathare, Soumitra and De Choudhury, Munmun and Kumar, Neha},
  booktitle={Proceedings of the 2022 CHI Conference on Human Factors in Computing Systems},
  pages={1--23},
  year={2022}
}

@article{bossewitch2022digital,
  title={Digital Futures in Mind: Reflecting on Technological Experiments in Mental Health \& Crisis Support},
  author={Bossewitch, Jonah and Brown, Lydia XZ and Gooding, Piers M and Harris, Leah and Horton, James and Katterl, Simon and Myrick, Keris and Ubozoh, Kelechi and Vasquez Encalada, Alberto},
  journal={Available at SSRN 4215994},
  year={2022}
}

@misc{replika2023,
  title = {Replika - Virtual AI Friend},
  howpublished = {\url{https://apps.apple.com/us/app/replika-virtual-ai-friend/id1158555867}},
  year = {2023},
  note = {[Internet]. App Store. [cited 2023 Mar 14]},
}

@inproceedings{crasto2021,
  author = {Crasto, R. and Dias, L. and Miranda, D. and Kayande, D.},
  title = {CareBot: A Mental Health ChatBot},
  booktitle = {2021 2nd International Conference for Emerging Technology (INCET)},
  pages = {1--5},
  year = {2021},
  address = {Belagavi, India},
  publisher = {IEEE},
  howpublished = {\url{https://ieeexplore.ieee.org/document/9456326/}},
  note = {[Internet]. [cited 2023 Mar 14]},
}

@article{oleary2023,
  author = {O'Leary, K.},
  title = {Human–AI collaboration boosts mental health support},
  journal = {Nat Med},
  year = {2023},
  url = {https://www.nature.com/articles/d41591-023-00022-w}
}

@misc{xiangsuicide2023,
  author = {Xiang, Chloe},
  title = {'He Would Still Be Here': Man Dies by Suicide After Talking with AI Chatbot, Widow Says},
  year = {2023},
  publisher = {Vice},
  url = {https://www.vice.com/en/article/pkadgm/man-dies-by-suicide-after-talking-with-ai-chatbot-widow-says}
}

@misc{xiangtessa2023,
  author = {Xiang, Chloe},
  title = {Eating Disorder Helpline Fires Staff, Transitions to Chatbot After Unionization},
  year = {2023},
  publisher = {Vice},
  url = {https://www.vice.com/en/article/n7ezkm/eating-disorder-helpline-fires-staff-transitions-to-chatbot-after-unionization}
}

@article{jargon2023,
  author = {Jargon, Julie},
  title = {How a Chatbot Went Rogue},
  journal = {Wall Street Journal},
  year = {2023},
  url = {https://www.wsj.com/articles/how-a-chatbot-went-rogue-431ff9f9}
}

@misc{Glorioso2023,
  author = {Glorioso, Chris},
  title = {Fake news? ChatGPT has a knack for making up phony anonymous sources},
  howpublished = {\url{https://www.nbcnewyork.com/investigations/fake-news-chatgpt-has-a-knack-for-making-up-phony-anonymous-sources/4120307/}},
  year = {2023},
  note = {Accessed: 2023-11-03},
  journal = {NBC New York}
}

@online{chiang2023,
  title={ChatGPT Is a Blurry JPEG of the Web},
  author={Chiang, Ted},
  year={2023},
  url={https://www.newyorker.com/tech/annals-of-technology/chatgpt-is-a-blurry-jpeg-of-the-web},
  journal={New Yorker},
}

@misc{HsuThompson2023,
  author = {Hsu, Tiffany. and Thompson, Stuart. A.},
  title = {Disinformation researchers raise alarms about A.I. Chatbots},
  howpublished = {The New York Times},
  year = {2023}
}

@misc{Klepper2023,
  author = {Klepper, David},
  title = {It turns out that ChatGPT is really good at creating online propaganda: 'I think what's clear is that in the wrong hands there's going to be a lot of trouble'},
  howpublished = {Fortune},
  year = {2023}
}

@inproceedings{bender2021dangers,
  title={On the dangers of stochastic parrots: Can language models be too big?},
  author={Bender, Emily M and Gebru, Timnit and McMillan-Major, Angelina and Shmitchell, Shmargaret},
  booktitle={Proceedings of the 2021 ACM conference on fairness, accountability, and transparency},
  pages={610--623},
  year={2021}
}

@article{zack2023coding,
  title={Coding Inequity: Assessing GPT-4's Potential for Perpetuating Racial and Gender Biases in Healthcare},
  author={Zack, Travis and Lehman, Eric and Suzgun, Mirac and Rodriguez, Jorge A and Celi, Leo Anthony and Gichoya, Judy and Jurafsky, Dan and Szolovits, Peter and Bates, David W and Abdulnour, Raja-Elie E and others},
  journal={medRxiv},
  pages={2023--07},
  year={2023},
  publisher={Cold Spring Harbor Laboratory Press}
}

@article{ernala2017linguistic,
  title={Linguistic markers indicating therapeutic outcomes of social media disclosures of schizophrenia},
  author={Ernala, Sindhu Kiranmai and Rizvi, Asra F and Birnbaum, Michael L and Kane, John M and De Choudhury, Munmun},
  journal={Proceedings of the ACM on Human-Computer Interaction},
  volume={1},
  number={CSCW},
  pages={1--27},
  year={2017},
  publisher={ACM New York, NY, USA}
}

@article{yoo2024missed,
  title={Missed Opportunities for Human-Centered AI Research: Understanding Stakeholder Collaboration in Mental Health AI Research},
  author={Yoo, Dong Whi and Woo, Hayoung and Pendse, Sachin R. and Lu, Nathaniel and Birnbaum, Michael L. and Abowd, Gregory and De Choudhury, Munmun},
  journal={Proceedings of the ACM on Human-Computer Interaction},
  volume={},
  number={CSCW},
  year={2024},
  publisher={ACM New York, NY, USA}
}

@inproceedings{li2023dimensions,
  title={The Dimensions of Data Labor: A Road Map for Researchers, Activists, and Policymakers to Empower Data Producers},
  author={Li, Hanlin and Vincent, Nicholas and Chancellor, Stevie and Hecht, Brent},
  booktitle={Proceedings of the 2023 ACM Conference on Fairness, Accountability, and Transparency},
  pages={1151--1161},
  year={2023}
}

@inproceedings{de2013predictingPPDChanges,
  title={Predicting postpartum changes in emotion and behavior via social media},
  author={De Choudhury, Munmun and Counts, Scott and Horvitz, Eric},
  booktitle={Proceedings of the SIGCHI conference on human factors in computing systems},
  pages={3267--3276},
  year={2013}
}

@article{chancellor2020methods,
  title={Methods in predictive techniques for mental health status on social media: a critical review},
  author={Chancellor, Stevie and De Choudhury, Munmun},
  journal={NPJ digital medicine},
  volume={3},
  number={1},
  pages={43},
  year={2020},
  publisher={Nature Publishing Group UK London}
}

@article{moura2022digital,
  title={Digital phenotyping of mental health using multimodal sensing of multiple situations of interest: A systematic literature review},
  author={Moura, Ivan and Teles, Ariel and Viana, Davi and Marques, Jean and Coutinho, Luciano and Silva, Francisco},
  journal={Journal of Biomedical Informatics},
  pages={104278},
  year={2022},
  publisher={Elsevier}
}

@inproceedings{de2016discovering,
  title={Discovering shifts to suicidal ideation from mental health content in social media},
  author={De Choudhury, Munmun and Kiciman, Emre and Dredze, Mark and Coppersmith, Glen and Kumar, Mrinal},
  booktitle={Proceedings of the 2016 CHI Conference on Human Factors in Computing Systems},
  pages={2098--2110},
  year={2016},
  organization={ACM}
}

@inproceedings{de2013predicting,
  title={Predicting depression via social media},
  author={De Choudhury, Munmun and Gamon, Michael and Counts, Scott and Horvitz, Eric},
  booktitle={ICWSM},
  year={2013}
}

@article{chancellor2019human,
  title={Who is the "Human" in Human-Centered Machine Learning: The Case of Predicting Mental Health from Social Media},
  author={Chancellor, Stevie and Baumer, Eric PS and De Choudhury, Munmun},
  year={2019},
  journal={PACM HCI},
    number={CSCW}
}

@inproceedings{sharma2018mental,
  title={Mental Health Support and its Relationship to Linguistic Accommodation in Online Communities},
  author={Sharma, Eva and De Choudhury, Munmun},
  booktitle={CHI},
  year={2018}
}

@inproceedings{chancellor2019opioid,
  author = {Chancellor, Stevie and Nitzburg, George and Hu, Andrea and Zampieri, Francesco and De Choudhury, Munmun},
 title = {Discovering Alternative Treatments for Opioid Use Recovery Using Social Media},
 booktitle = {Proc. CHI},
 year = {2019}
}


\end{document}